\newif\ifpictures
\picturestrue

\newif\ifcomment
\commenttrue

\documentclass[11pt]{amsart}

\usepackage{latex_base}
\usepackage{macros}
\usepackage[table]{xcolor}
\usepackage[dvipsnames]{xcolor}
\usepackage{tablefootnote}
\usepackage{wrapfig}
\usepackage{caption}
\usepackage{afterpage}
\usepackage{booktabs}
\usepackage{textcomp}

\author[J. Naumann]{Jonas Naumann}
\address{Jonas Naumann, German Aerospace Center (DLR), Institute of Lightweight Systems, Composite Process Technologies, Ottenbecker Damm 12, 21682 Stade, Germany\medskip}
\email{j.naumann@dlr.de}

\author[J. P. Appels]{Jonas P. Appels}
\address{Jonas P. Appels, German Aerospace Center (DLR), Institute of Lightweight Systems, Composite Process Technologies, Ottenbecker Damm 12, 21682 Stade, Germany\medskip}
\email{jonas.appels@dlr.de}

\author[J. Biermann]{Julius Biermann}
\address{Julius Biermann, German Aerospace Center (DLR), Institute of Lightweight Systems, Composite Process Technologies, Ottenbecker Damm 12, 21682 Stade, Germany\medskip}
\email{julius.biermann@dlr.de}

\author[C. Gorsky]{Christopher Gorsky}
\address{Christopher Gorsky, German Aerospace Center (DLR), Institute of Lightweight Systems, Structural Mechanics, Lilienthalplatz 7, 38108 Braunschweig, Germany\medskip}
\email{christopher.gorsky@dlr.de}

\author[T. de Wolff]{Timo de Wolff$^\ast$}
\address{Timo de Wolff, Technische Universit\"at Braunschweig, Institute of Analysis and Algebra, Applied Algebra, Universit\"atsplatz 2, 38106 Braunschweig, Germany\medskip}
\email{t.de-wolff@tu-braunschweig.de}
\thanks{$^\ast$Corresponding authors.}

\author[C. Brauer]{Christoph Brauer$^\ast$}
\address{Christoph Brauer, German Aerospace Center (DLR), Institute of Lightweight Systems, Composite Process Technologies, Ottenbecker Damm 12, 21682 Stade, Germany\medskip}
\email{christoph.brauer@dlr.de}

\title[Automatic Ply-specific Analyses of CFRP Micrographs]{Automatic Ply-specific Analyses of CFRP Micrographs using Shortest-Path-based Ply Distinction}

\keywords{ply detection, composite micrograph analysis, shortest path algorithm, automation, fiber volume fraction analysis, ply thickness measurement, interleaf thickness measurement}

\begin{document}

\begin{abstract}
We present an automated approach to distinguish between ply instances in semantic segmentation masks of high-resolution carbon-fiber reinforced polymer micrographs. Interpreting the segmentation mask as a graph with pixels as vertices, enables us to use a shortest-path algorithm yielding the ply-separating paths.
Thereby, we bridge the gap between semantic segmentation and ply instance segmentation using global information.  
We successfully apply our approach on high-resolution micrographs featuring a broad range of characteristics like artificially added gaps in single or multiple plies, different stacking sequences and ply traversing cracks.
Assigning each fiber pixel to a ply based on the calculated paths, allows for a comprehensive, quantitative ply analysis with respect to its microstructural properties like the local fiber volume fraction as well as locally resolved ply and interleaf layer thickness.  
These insights help to reveal manufacturing-induced inhomogeneities, draw conclusions on manufacturing parameters and link mechanical properties to underlying microstructural imperfections. 
\end{abstract}

\maketitle

\section{Introduction}

Carbon-fiber reinforced polymer (CFRP) composites are widely used in lightweight structures because they combine low density with high specific stiffness and strength, as well as good fatigue and corrosion resistance \cite{Alam.2019,Fu.2022}. At the same time, their macroscopic behavior is strongly governed by the underlying microstructure, including local fiber packing, ply geometry and interfacial morphology \cite{Katuin.2021}. Consequently, manufacturing-related inhomogeneities such as voids, resin-rich regions, fiber waviness and ply-thickness variations must be avoided or at least quantified, since they can impact structural performance \cite{Fu.2022}. Accordingly, the visual inspection of polished cross-sections by high-resolution microscopy remains a standard practice for evaluating laminate quality and detecting microstructural defects during the development phase but also for quality assurance processes during ongoing production \cite{Katuin.2021}.\newline

\begin{figure}[p]
\captionsetup{margin={0.4cm, 0.4cm}}
  \vspace{\baselineskip}
  \centering
\includegraphics[width=0.68\textwidth]{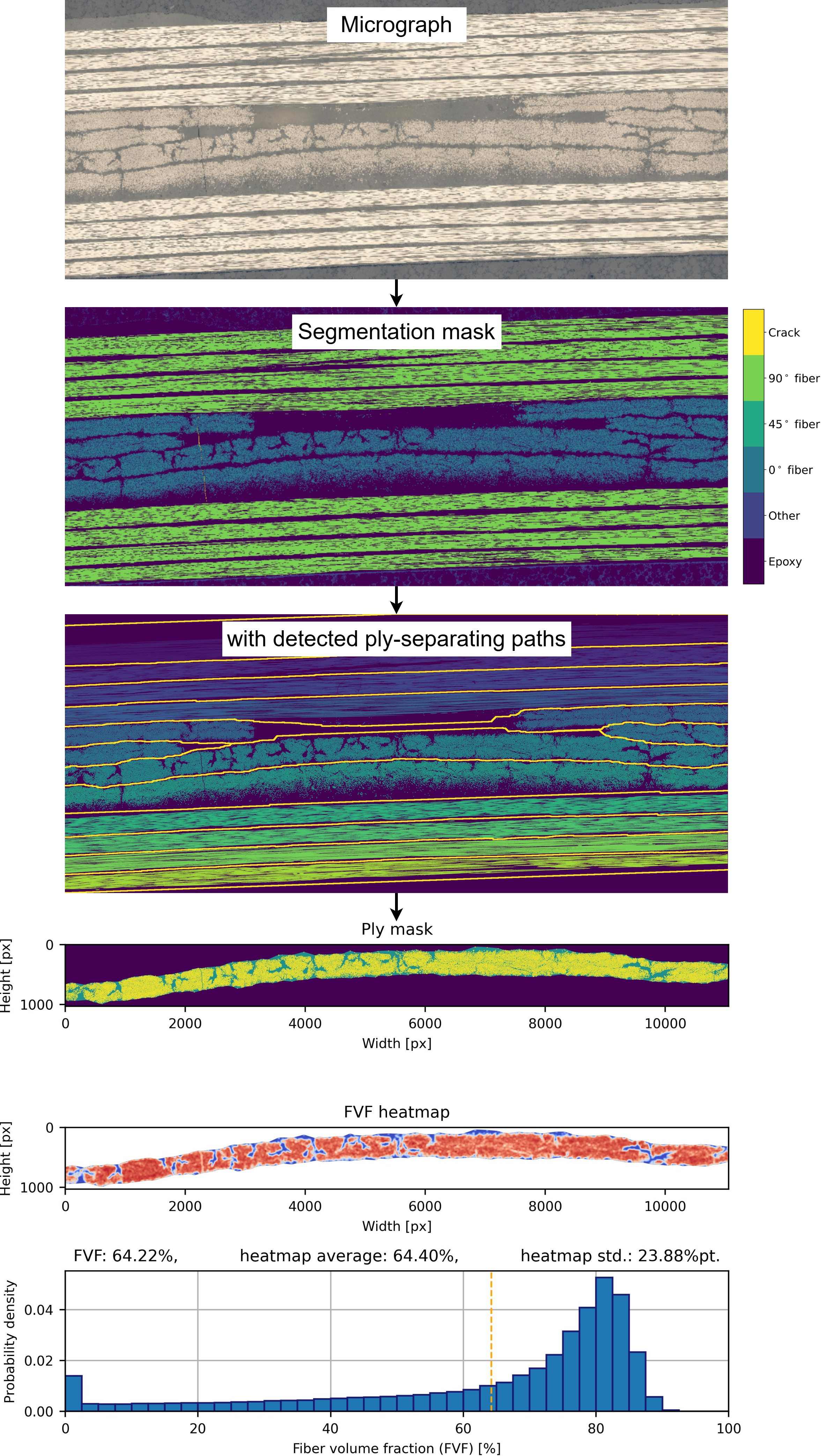}
    \caption{Visualized analysis pipeline from micrograph to local FVF analysis for ply instance. This paper is about the second half of this process.}
    \label{fig:Process}
\end{figure}

Segmentation algorithms can not only accelerate the subsequent, largely manual inspection process, but also enable the quantitative assessment of microstructural features that would otherwise remain qualitatively described. 
Subsequent to the application of our \struc{semantic segmentation} model \cite{Naumann.2025} trained from scratch on our own micrographs and suitable for crack detection and fiber volume fraction (FVF) analyses \cite{Appels.2025}, we aim to distinguish between fiber ply instances. Automating the assignment of every single fiber to its ply as one would intuitively decide based on their locations and clusters would enable us to efficiently analyze and compare separate plies regarding their thickness and fiber distribution.\newline

Lau, Belnoue, and Hallett (2023) \cite{Lau.2023} propose an approach for ply identification in composite laminates based on relatively low resolution micrographs with an average pixel size of 14.5 µm. It detects the center of each ply based on brightness values separately for image column slices and connects them based on the shortest distance to the centers of neighboring slices. Ply borders are finally derived as midpoint between two ply centers. 
The algorithm is suitable for different levels of waviness, is limited to plies of same length, but is not applicable to tapered or curved laminates. Potential use of the extracted geometries includes 3D finite element method for stress and strength simulations as well as investigation of geometry deviations between \textit{as manufactured} and \textit{as designed}.
In contrast to \cite{Lau.2023}, our method is focused on high resolution micrographs with a pixel size between 0.4 and 0.5 µm due to the downstream analyses. Therefore, we require a more fine-grained result to assign every single fiber to a ply instance.\newline

Our main contribution is an approach to obtain \struc{ply-separating paths} for CFRP segmentation masks based on a \struc{shortest path algorithm}. The general approach is broadly applicable and only presumes ply separability based on interleaf layers or different fiber orientation angles provided in the segmentation mask. It can handle interrupted or wavy plies and does not require user interaction except of few parameter choices. The usage of our machine learning-based segmentation model has advantages, however even a basic, binary segmentation mask based on Otsu's method \cite{Otsu.1979} is sufficient for specimens with interleaf layers. Based on the inherent ply instance identification, we can extend existing segmentation mask analyses to the ply level and automatically measure interleaf and ply thicknesses. We describe corresponding use cases for these downstream analyses. An exemplary process from micrograph to ply instance-specific FVF analysis is visualized in \cref{fig:Process}.\newline

In the remainder of this paper, we describe the used material and data (\cref{sec: data acquisition}) as well as our approach (\cref{sec: methodology}, present an evaluation on selected micrographs (\cref{sec: evaluation}), discuss limitations (\cref{sec: challenges}), present several use cases (\cref{sec: downstream analyses}) and finally conclude and give an outlook on potential future work (\cref{sec: conclusion}).

\section{Data and methodology}
\subsection{Data acquisition}
\label{sec: data acquisition}
\begin{figure}[htbp]
    \centering

    \begin{subfigure}{0.18\textwidth}
        \centering
        \fbox{%
            \includegraphics[width=\linewidth,height=3cm,keepaspectratio]{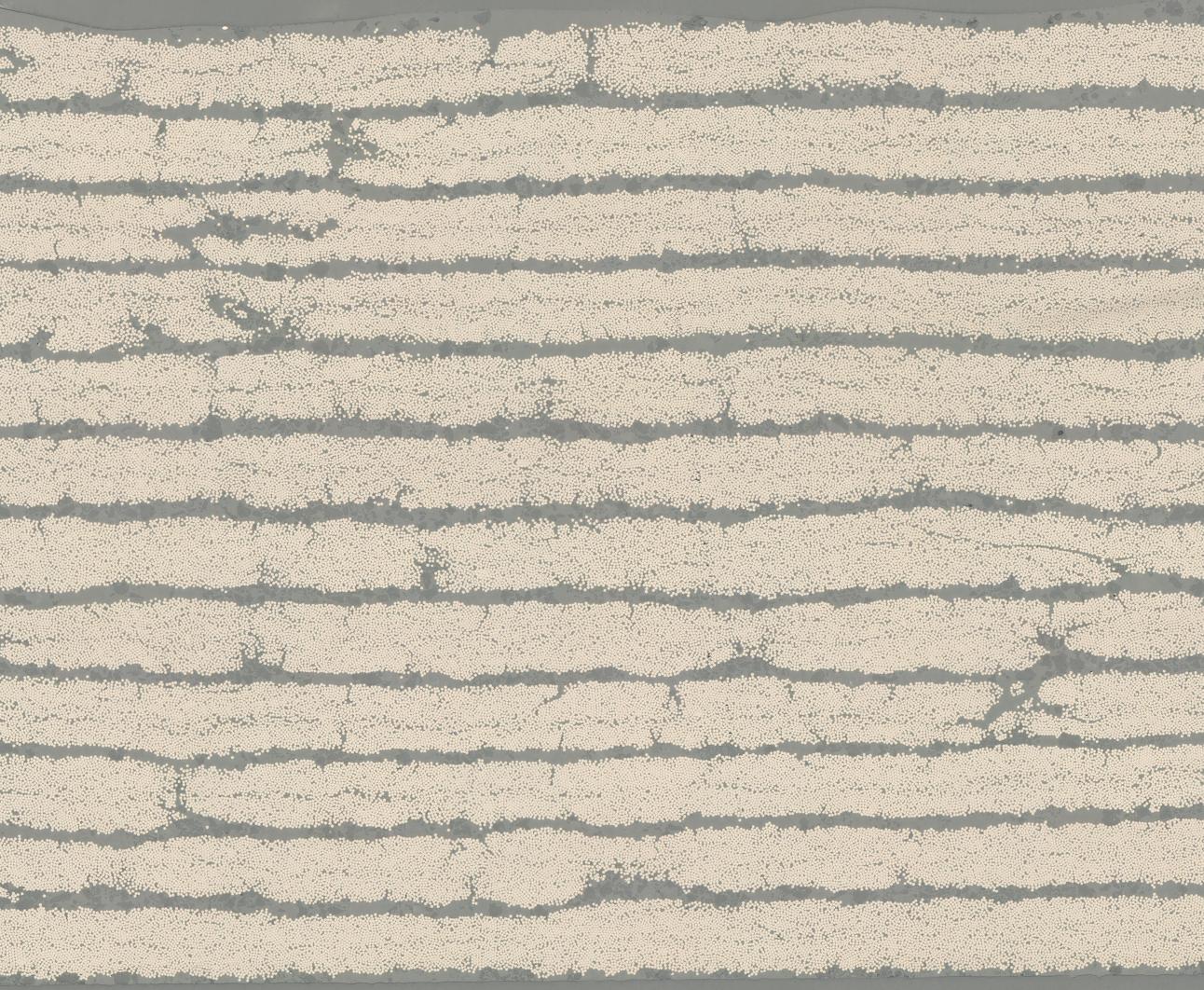}%
        }
        \caption{}
        \label{fig:A}
    \end{subfigure}
    \hfill
    \begin{subfigure}{0.18\textwidth}
        \centering
        \fbox{%
            \includegraphics[width=\linewidth,height=3cm,keepaspectratio]{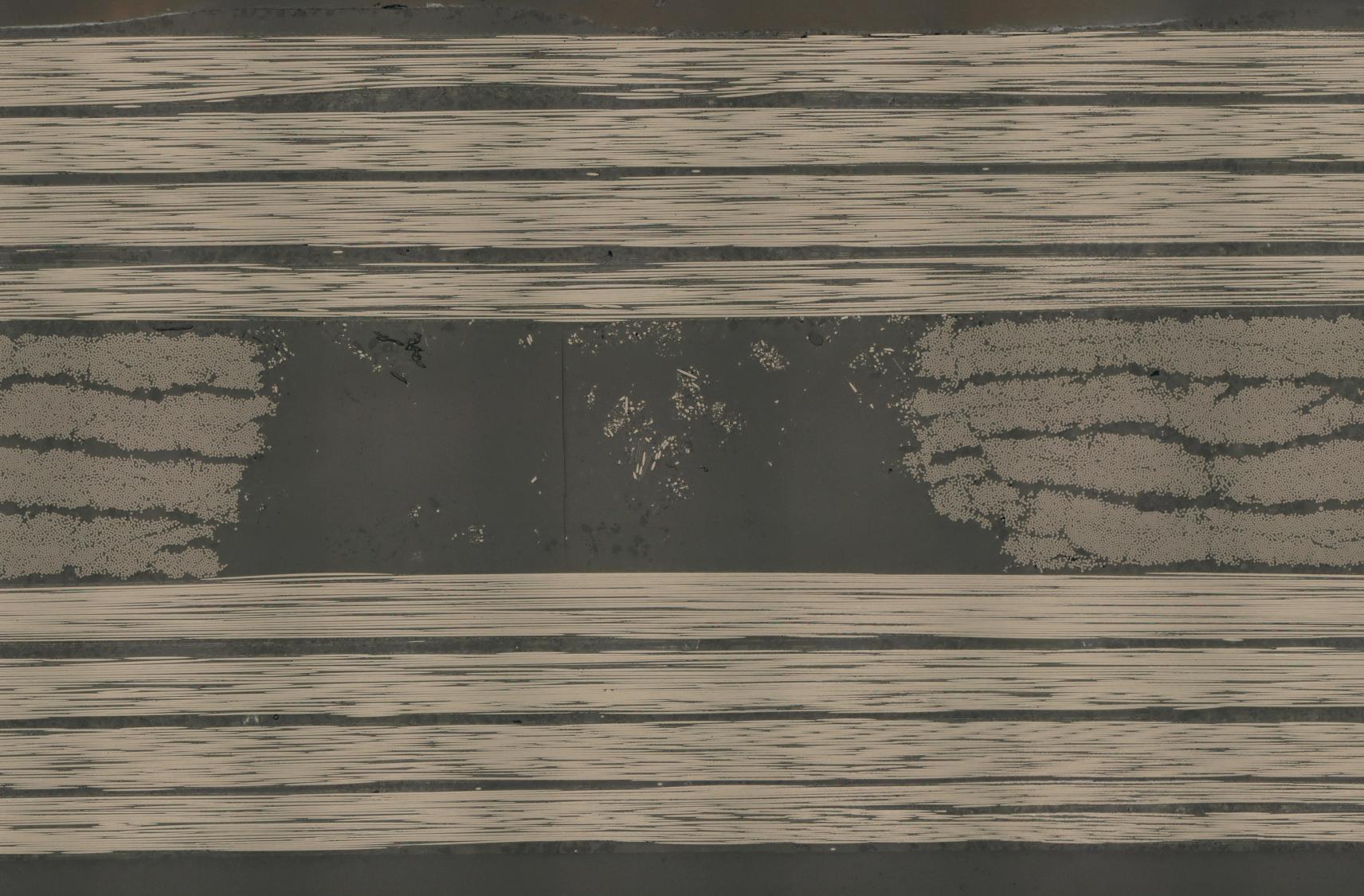}%
        }
        \caption{}
        \label{fig:B}
    \end{subfigure}
    \hfill
    \begin{subfigure}{0.18\textwidth}
        \centering
        \fbox{%
            \includegraphics[width=\linewidth,height=3cm,keepaspectratio]{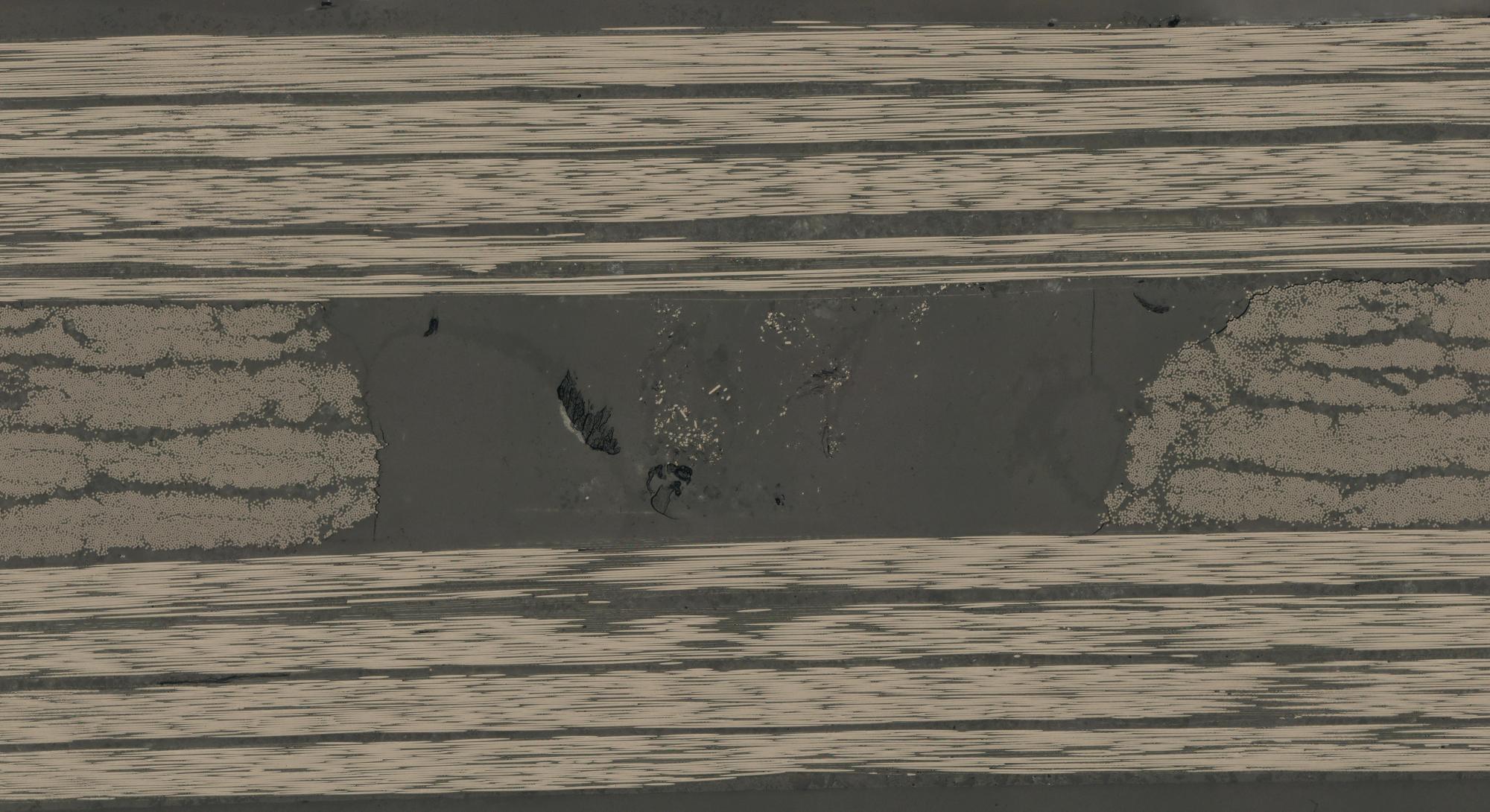}%
        }
        \caption{}
        \label{fig:C}
    \end{subfigure}
    \hfill
    \begin{subfigure}{0.18\textwidth}
        \centering
        \fbox{%
            \includegraphics[width=\linewidth,height=3cm,keepaspectratio]{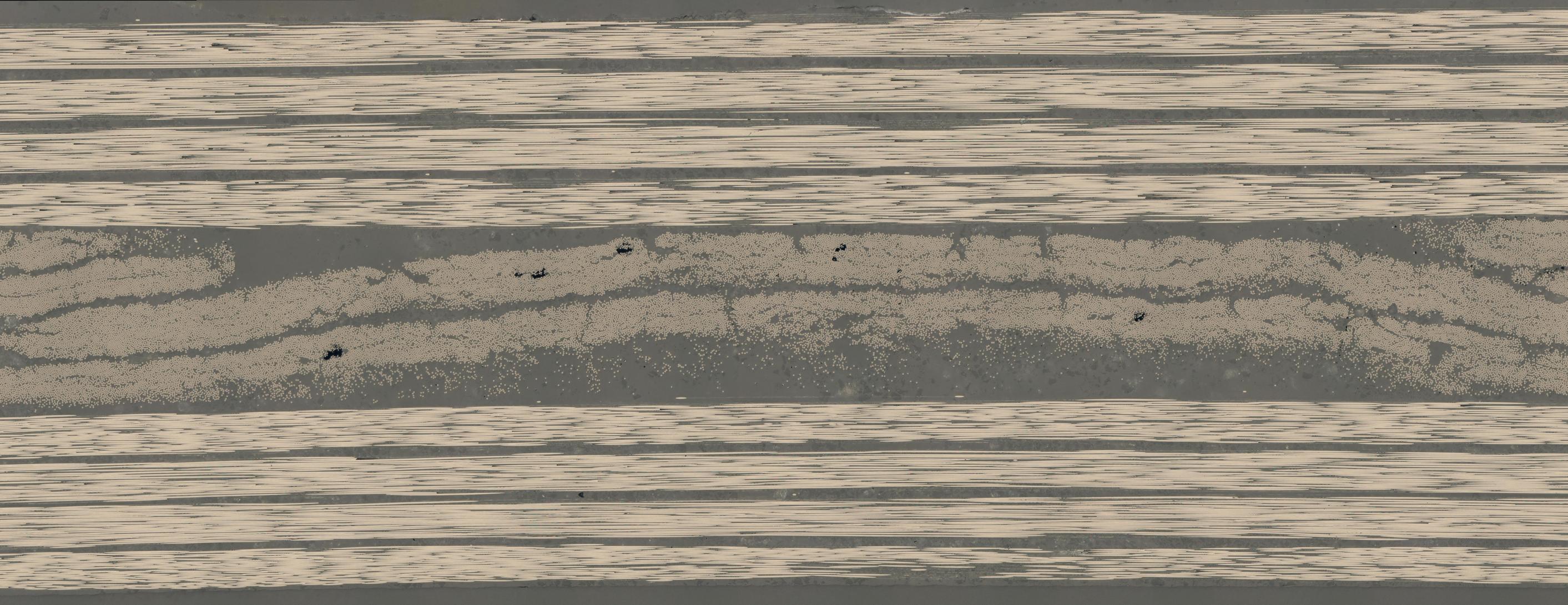}%
        }
        \caption{}
        \label{fig:D}
    \end{subfigure}
    \hfill
    \begin{subfigure}{0.18\textwidth}
        \centering
        \fbox{%
            \includegraphics[width=\linewidth,height=3cm,keepaspectratio]{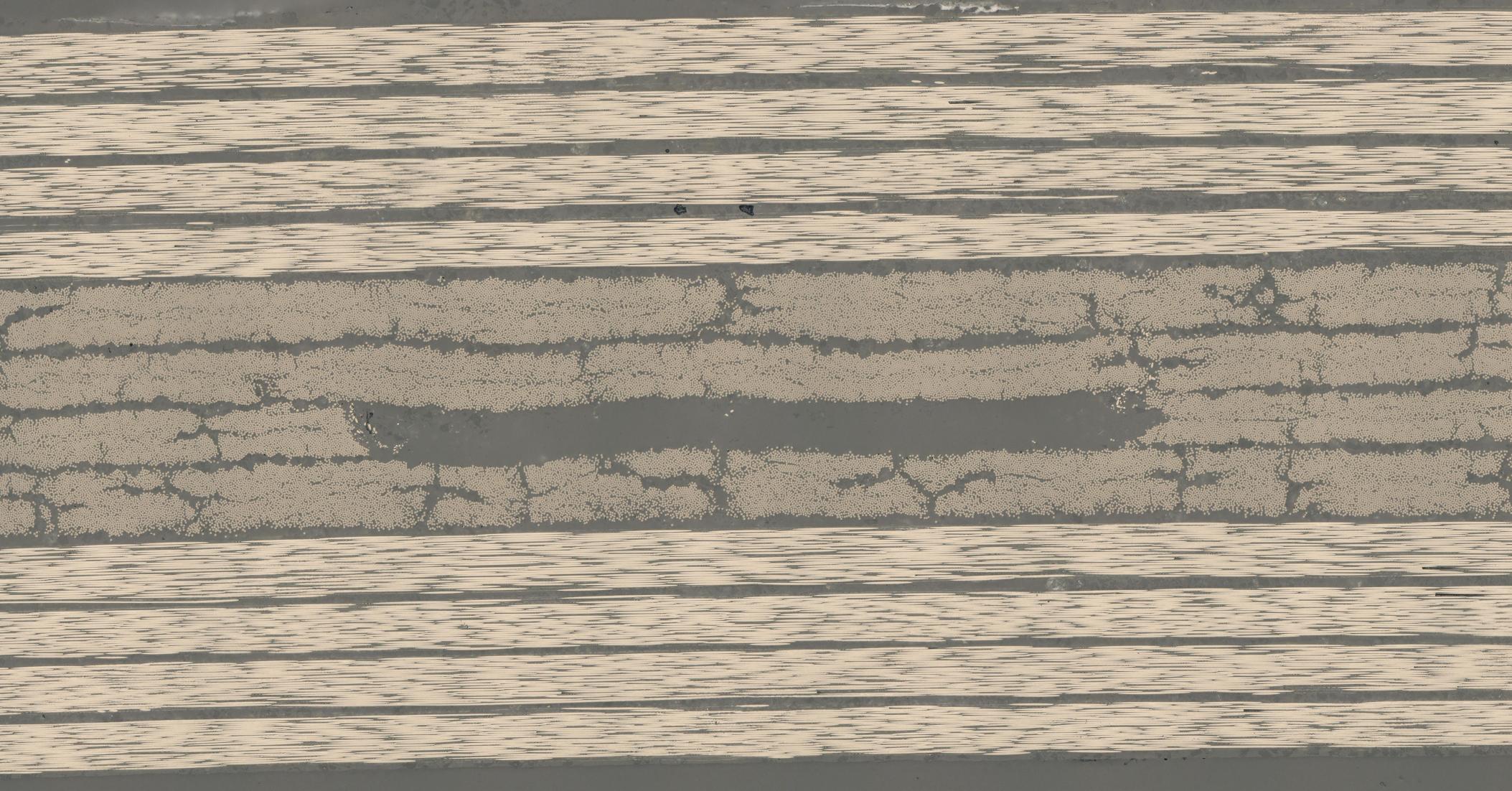}%
        }
        \caption{}
        \label{fig:E}
    \end{subfigure}

    \vspace{0.5cm}

    \begin{subfigure}{0.18\textwidth}
        \centering
        \fbox{%
            \includegraphics[width=\linewidth,height=3cm,keepaspectratio]{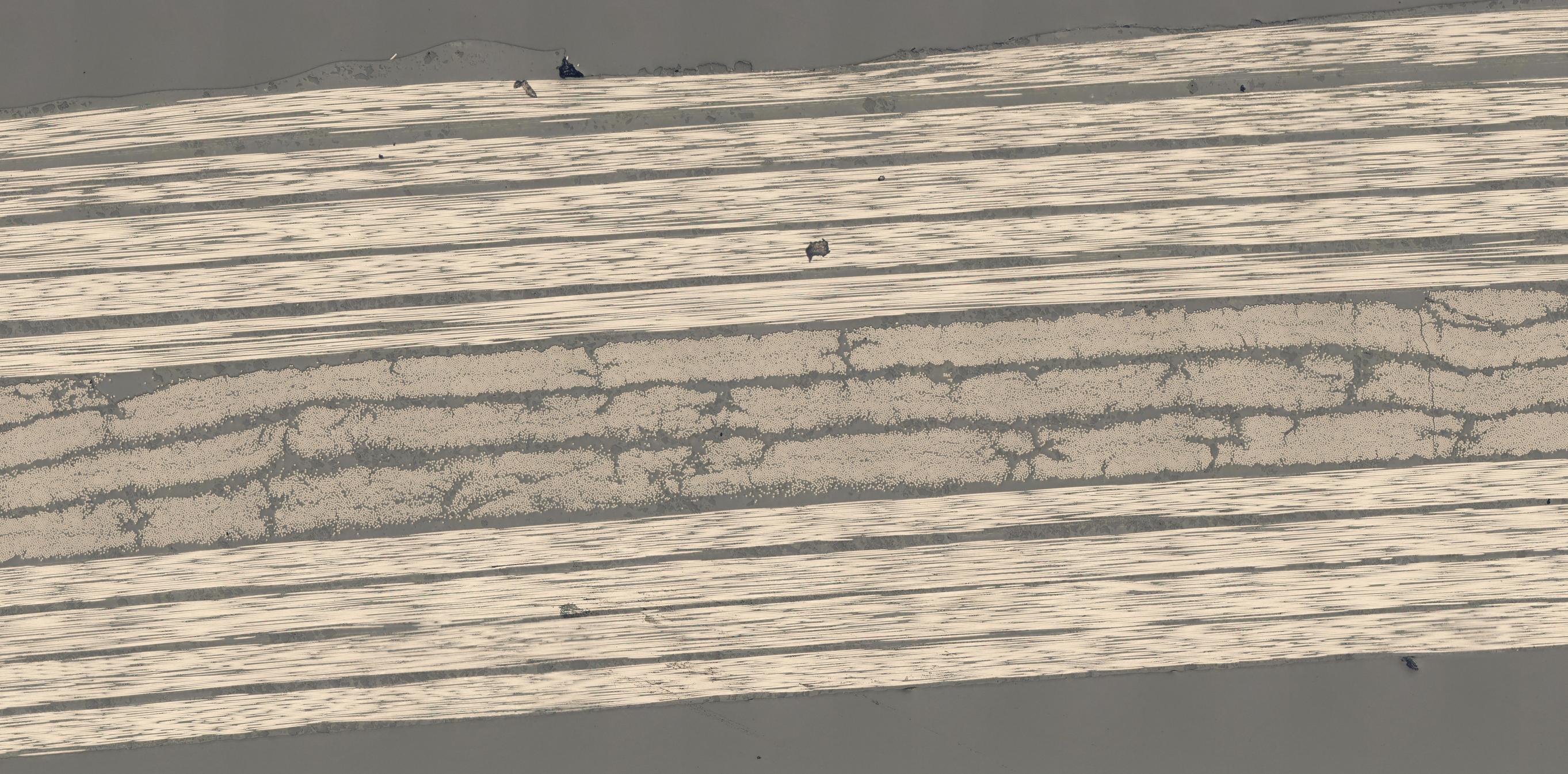}%
        }
        \caption{}
        \label{fig:F}
    \end{subfigure}
    \hfill
    \begin{subfigure}{0.18\textwidth}
        \centering
        \fbox{%
            \includegraphics[width=\linewidth,height=3cm,keepaspectratio]{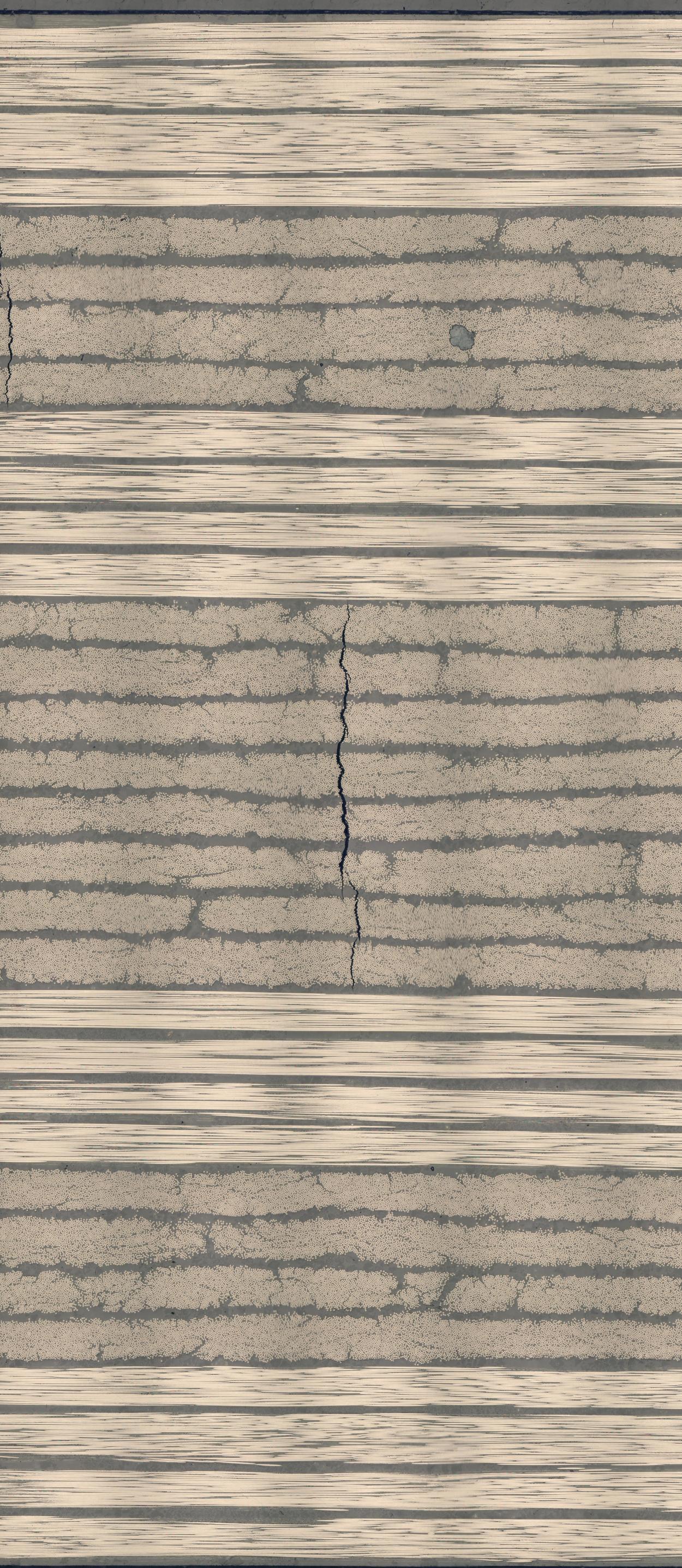}%
        }
        \caption{}
        \label{fig:G}
    \end{subfigure}
    \hfill
    \begin{subfigure}{0.18\textwidth}
        \centering
        \fbox{%
            \includegraphics[width=\linewidth,height=3cm,keepaspectratio]{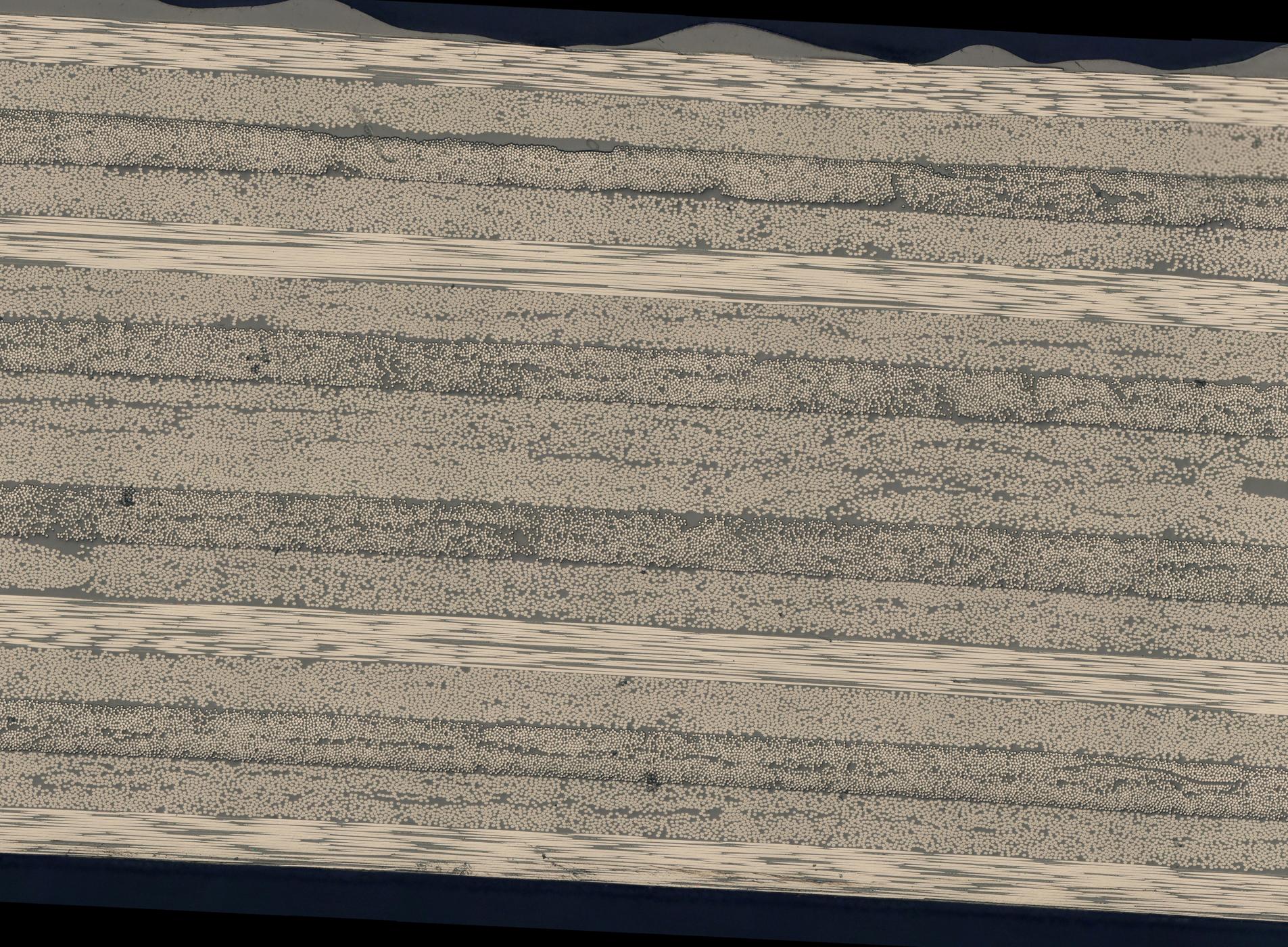}%
        }
        \caption{}
        \label{fig:H}
    \end{subfigure}
    \hfill
    \begin{subfigure}{0.18\textwidth}
        \centering
        \fbox{%
            \includegraphics[width=\linewidth,height=3cm,keepaspectratio]{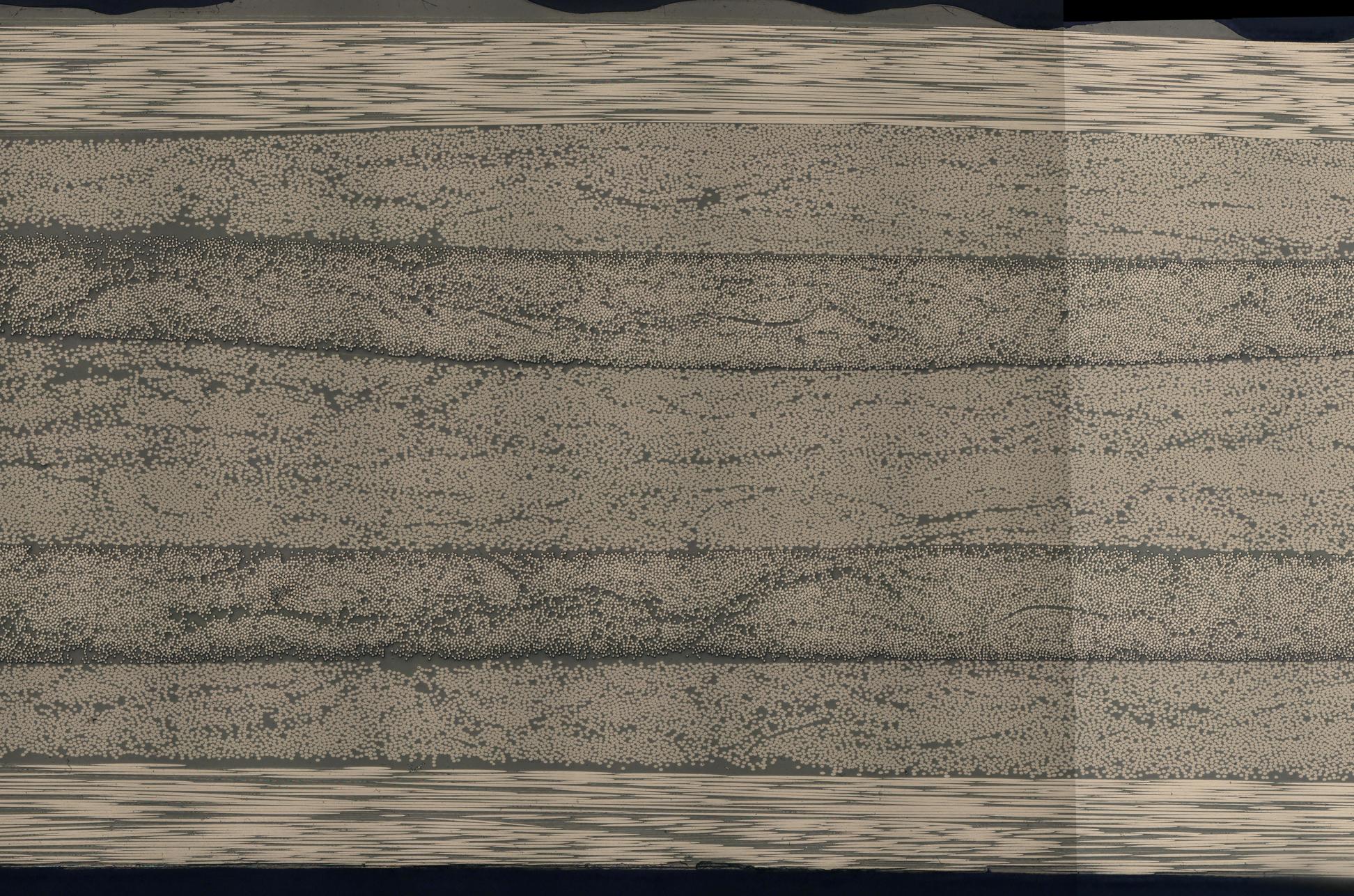}%
        }
        \caption{}
        \label{fig:I}
    \end{subfigure}
    \hfill
    \begin{subfigure}{0.18\textwidth}
        \centering
        \fbox{%
            \includegraphics[width=\linewidth,height=3cm,keepaspectratio]{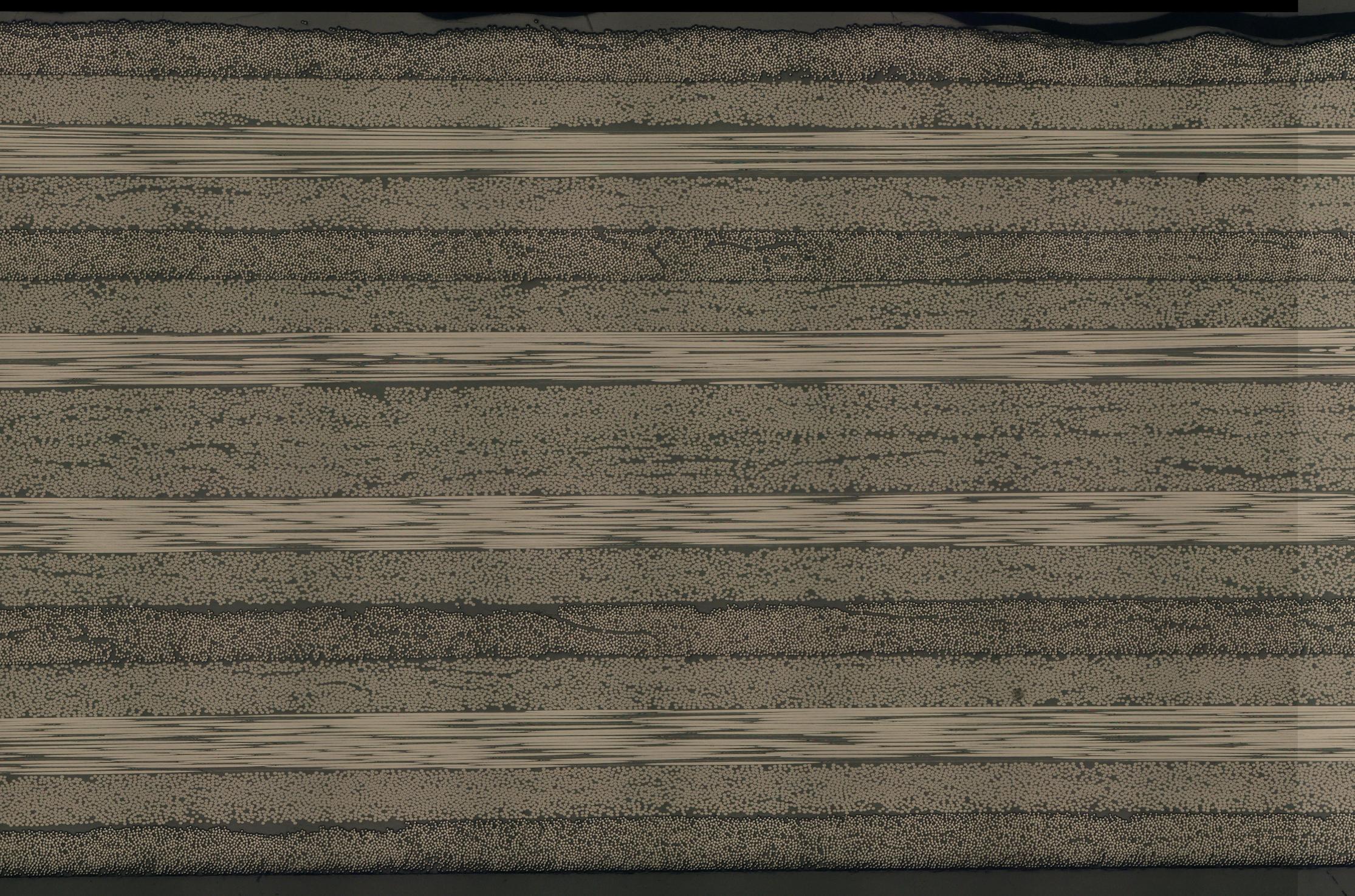}%
        }
        \caption{}
        \label{fig:J}
    \end{subfigure}

    \caption{All specimens used for evaluation [source: DLR].}
    \label{fig:overview_specimens}
\end{figure}

To ensure a robust ply recognition, we developed the algorithm using a variety of materials, layups, material conditions, and artificial artefacts. We show the micrographs used for the evaluation (see \cref{sec: evaluation}) in \cref{fig:overview_specimens} and provide corresponding material and manufacturing information in \cref{tab: material}. Note that the micrographs of \cref{fig:Process,fig: interleaf thickness analysis,fig: fiber ply thickness} are not part of that evaluation dataset presented here. \newline

The micrographs have the following special characteristics:
For Image (A), we observe a vertical gradient in FVF \cite{Appels.2025} and it is the only example here only featuring one fiber orientation angle. 
Images (B) and (C) show gaps across plies 5 to 8 (counted from bottom to top), while image (D) shows a gap in plies 7 and 8, accompanied by a pronounced upward displacement of plies 5 and 6. 
Image (E) contains a gap in ply 6 only. 
Image (F) shows a wide gap in ply 8 with upward displacement of the underlying plies. 
Specimen (G) was subjected to prolonged water exposure followed by thermal cycling between 77\,K and room temperature, resulting in the microcrack visible in the image. 
The images (H), (I), and (J) were acquired under more challenging lighting conditions, characterized by varying brightness. \newline

All of the micrographs are captured with a \textit{Keyence VHX-5000} digital microscope and have an approximate pixel size between 0.4 and 0.5 µm and the used excerpts shown in \cref{fig:overview_specimens} have heights and widths between 4,250 and 11,500 px.

\begin{center}
\begin{table}
\caption{Material and manufacturing information for evaluation specimens.}
\label{tab: material}
\centering
 \begin{tabular}{ ccccccc } 
\toprule
& && & Nominal &   \\ 
& Matrix && Stacking& areal & Nominal  \\ 
  Image & system & Fibers &sequence\tablefootnote{The values in the brackets denote the fiber angle per ply (from bottom to top), the subscript numbers their repetitions and subscript $S$ stands for a symmetric repetition of the corresponding bracket.}&  weight & FVF & Tooling  \\ 
 \midrule
  A& Teijin Q183 & IMS65 & $[0_{12}]$& 145\,gsm & 65\,\%  & curved \\
  B-E& Teijin Q183 & IMS65 & $[90_4/0_4/90_4]$& 145\,gsm & 65\,\%  & curved \\
  F& Teijin Q183 & IMS65 & $[90_4/0_4/90_5]$& 145\,gsm & 65\,\%  & curved \\
 G& Teijin Q183 &IMS65& $[90_4/0_4/90_4/0_4]_S$ & 145\,gsm & 65\,\%  & flat \\
 H& HexPly 6376 & HTS& $[90/45/0/$-$45/90/45/0/$-$45]_S$& 134\,gsm & 65\,\% &curved\\
 I& HexPly 6376 & HTS& $[90/45/0/$-$45]_S$& 268\,gsm & 65\,\% & curved  \\
 J& HexPly 6376 & HTS& $[0/45/90/$-$45/0/45/90/$-$45]_S$& 134 gsm & 65\,\% &curved\\
 \bottomrule
\end{tabular}
\end{table}
\end{center}

We derive the corresponding segmentation masks by applying our semantic segmentation model \cite{Naumann.2025} based on the U-Net \cite{Ronneberger.2015} and InternImage backbone \cite{Wang.2023}. This model has been trained from scratch on our own, labeled dataset to distinguish between epoxy, 0$^{\circ}$ fiber, 45$^{\circ}$ fiber, 90$^{\circ}$ fiber (as-designed angles), crack and other (e.g. dirt). To reduce the barriers to use our approach, we also test it on masks derived by basic, binary (fibers vs. matrix) Otsu's method \cite{Otsu.1979} which thresholds greyscale values maximizing between-class and minimizing within-class variance.

\subsection{From segmentation mask to ply-separating paths}
\label{sec: methodology}
Our approach is based on using a \struc{shortest path algorithm} to find the best ply-separating paths through the segmentation mask. 
In short, it consists of five steps: 
\begin{enumerate}
    \item We assign costs to every single pixel of the segmentation mask based on its minimum distances to specific classes (\cref{sec: costs to pixels}).
    \item We calculate start points on the left border based on local costs (\cref{sec: start point calc}).
    \item We derive a graph from the cost mask where each pixel corresponds to one vertex and the costs are assigned to the edges between them (\cref{sec: mask to graph}).
    \item We apply Dijkstra's algorithm to determine the shortest paths from all start points to all points on the right border.
    \item We make a preselection of end locations on the right side and perform a bipartite matching of feasible combinations of start and end locations to find the cheapest combination of the required number of paths (\cref{sec: end point and path selection}). 
\end{enumerate}

\subsubsection{Assigning costs to pixels}
\label{sec: costs to pixels}
We assign a cost to every pixel taking into account (1) the pixel's minimum distance to any fiber pixel, (2) its minimum distance to any non-fiber pixel, and (3) - if applicable for the specimen and segmentation mask at hand - we reward spatial proximity to fibers of different orientation, i.e., having a small distance to two different fiber classes. \newline

Let \struc{$M$$\in \{1,2,3,4,5,6\}^{H\times W}$} be the six-class \struc{segmentation mask} with height and width $H,W\in \mathbb{N}$. The corresponding \struc{cost mask} \struc{$C\in \mathbb{R}^{H\times W}$} of same size is calculated as follows: 
\begin{align}
\notag & C_{h,w}&=&\underbrace{w_1C_{1_{h,w}}}_{\textrm{\parbox{1in}{\centering penalize distance to resin}}}+\underbrace{w_2C_{2_{h,w}}}_{\textrm{\parbox{1in}{\centering reward distance to fibers}}}+\underbrace{w_3C_{3_{h,w}}}_{\textrm{\parbox{1in}{\centering reward two close fiber angles}}}\\
\notag &&&\underbrace{-\min_{h,w}\left(w_1C_{1_{h,w}}+w_2C_{2_{h,w}}+w_3C_{3_{h,w}}\right)}_{\textrm{\parbox{1.2in}{\centering set lower bound to zero}}}\underbrace{+0.5}_{\textrm{\parbox{1in}{\centering ensure no free paths}}} 
\end{align}

where $w_1, w_2, w_3 \in \mathbb{R}_{\geq0}$ and 
\begin{align}
\notag &C_{1_{h,w}}&=\quad&\frac{\min(EDT_{\text{resin}}(M)_{h,w}, r_{\text{fiber}})}{r_{\text{fiber}}}\\ 
\notag &C_{2_{h,w}}&=\quad&-\sqrt{\frac{\max(\min(\left(EDT_{\text{fiber}}(M)-2\right)_{h,w}, \frac{t_{\text{interleaf}}}{2}), 0)}{\frac{t_{\text{interleaf}}}{2}}} \\
\notag &C_{3_{h,w}}&=\quad&\frac{\min(EDT_{\text{transition}}(M)_{h,w}, \frac{t_{\text{interleaf}}}{2})}{\frac{t_{\text{interleaf}}}{2}}-1 \\ \notag
\end{align}

\begin{figure}[h]
\captionsetup{margin={0.2cm, 0.2cm}}
  \centering
    \includegraphics[width=0.75\textwidth]{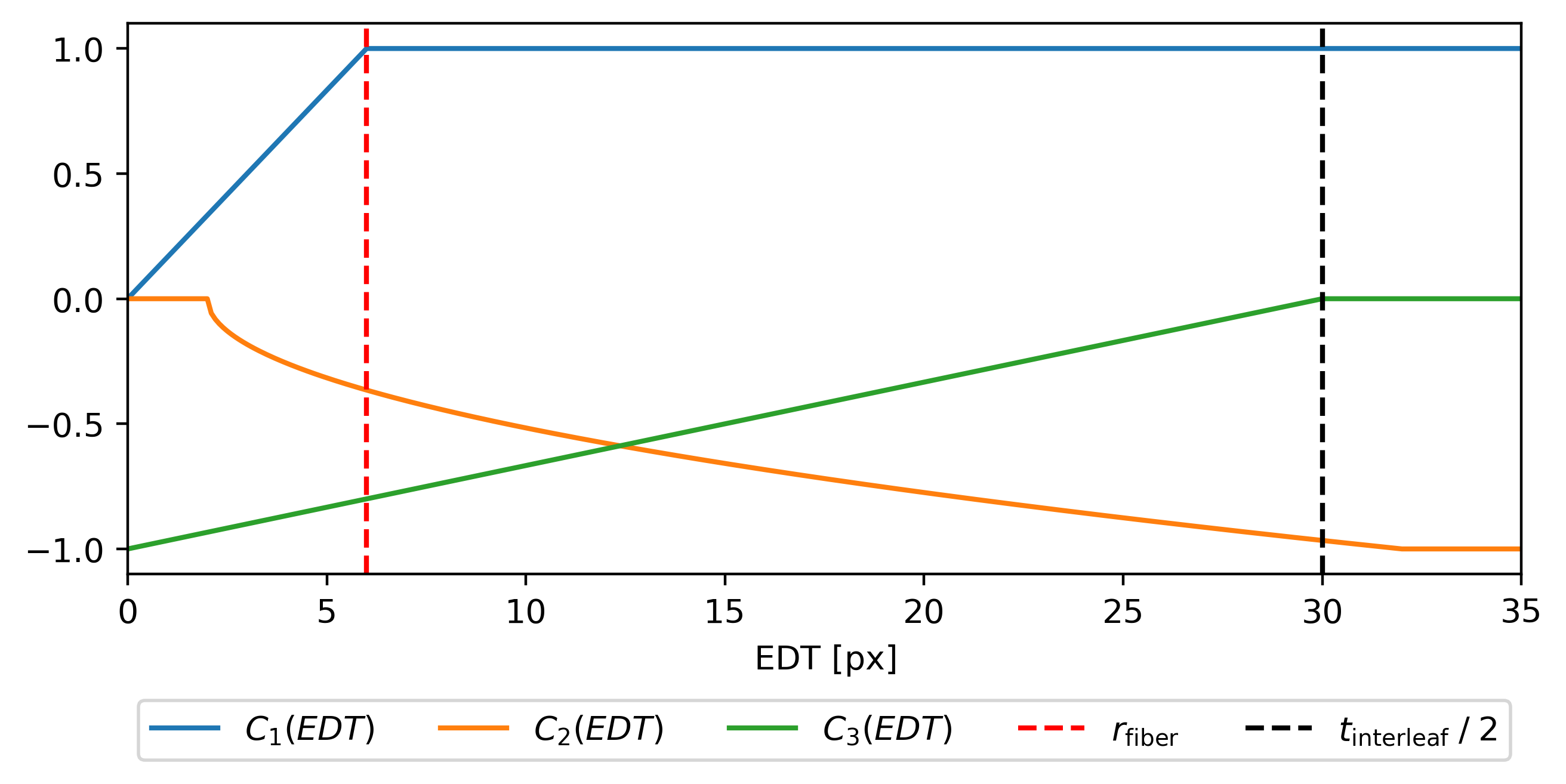}
    \caption{Visualization of all three cost terms $C_1$, $C_2$, and $C_3$ as a function of the respective $EDT$.}
    \label{Fig: cost_terms_plot}
\end{figure}

\begin{figure}[h]
\captionsetup{margin={0.2cm, 0.2cm}}
  \centering
    \includegraphics[width=0.8\textwidth]{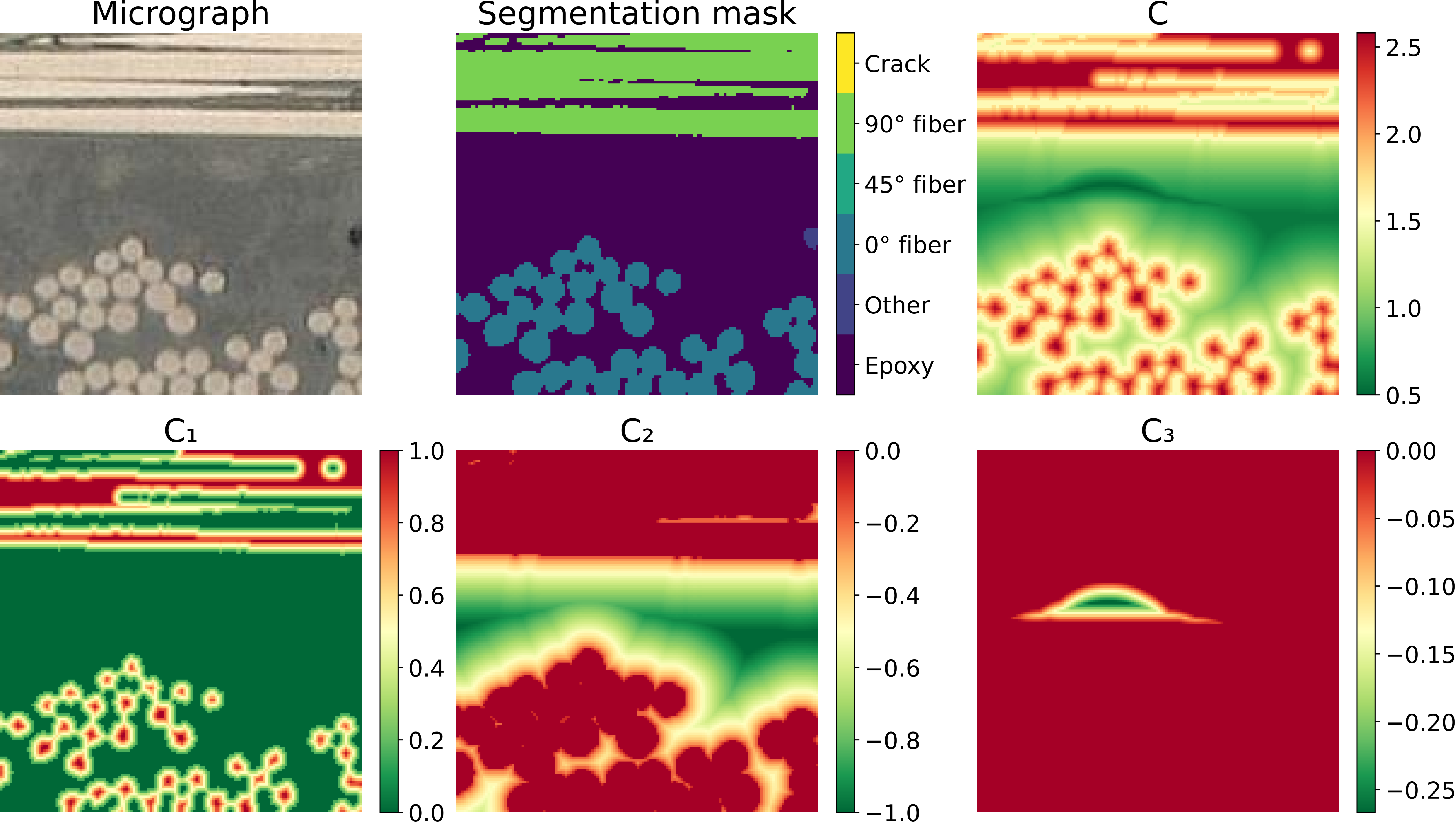}
    \caption{Visualization of all three cost terms $C_1, C_2, C_3$ as well as corresponding $C$ (for $w_1=w_2=w_3=1.0$), micrograph, and segmentation mask.}
    \label{Fig: cost_terms_example}
\end{figure}

Here, $EDT$ stands for \struc{Euclidean distance transform} and therefore $EDT_{c}(M)_{h,w}$ denotes the Euclidean distance of the pixel at height $h$ and width $w$ in $M$ to the closest pixel of class $c$.
Moreover, $EDT_{\text{transition}}(M)$ denotes the pixel-wise median of $EDT_{\text{0}^{\circ} \text{fiber}}(M)$, $EDT_{\text{45}^{\circ} \text{fiber}}(M)$, and $EDT_{\text{90}^{\circ} \text{fiber}}(M)$ which corresponds to the minimum distance to the second furthest fiber class and yields low values in transition areas from two plies of different fiber orientation. $EDT_{\text{fiber}}(M)$ considers the three fiber classes as one joint class. And last but not least, $r_{\text{fiber}}$ denotes the approximate fiber radius (in our case: $r_{\text{fiber}}=6$) and $t_{\text{interleaf}}$ a rough estimate of the average interleaf layer thickness (in our case: $\frac{t_{\text{interleaf}}}{2}=30$), both in pixels. We take the square root of the distance to fibers as we assume a decreasing benefit of increasing the path's distance to fiber pixels and only reward distances to fiber pixels larger than two. We visualize $C_1$, $C_2$, and $C_3$ as a function of the respective $EDT$ in \cref{Fig: cost_terms_plot} and together with $C$ for a small example in \cref{Fig: cost_terms_example}. 

For everything that follows, we downsample the cost mask by a factor of four for reasons of efficiency. In detail, we only keep every fourth pixel, which already contains information about the now omitted neighborhood, as the EDT is applied to the original resolution. To account for potential strong inclinations, we add a rotation angle as parameter. Before the graph is created (\cref{sec: mask to graph}), we apply this rotation to the pixel-wise cost mask as well as to the start points and undo it afterwards regarding the path locations.

Moreover, we create an additional cost mask only for the left border region used for the start point calculation described in the following \cref{sec: start point calc}. This allows for using different cost term weights for the start point calculation than for the path calculation. More details are provided in \cref{sec: challenges}.

\subsubsection{Start point calculation}
\label{sec: start point calc}
Next, we calculate start points on the left side of the mask, considering only a small border slice of the cost mask. In detail, each vertical position is assigned the average cost of its row's pixels\footnote{or of the sheared, formerly horizontal line of pixels in case of a nonzero rotation angle} and we iteratively add the cheapest vertical position as long as it maintains a set minimum distance to the already selected start points. Moreover, the two areas between a start point candidate and its already selected neighbors must fulfill a set minimum FVF to avoid selecting two start points in the same resin area between two plies. Since this calculation is based on local information, incorrect start locations can be selected and near, correct ones are then missed due to the distance threshold (see \cref{Fig: challenge 2} and \cref{sec: challenges} for an example). To prevent this, we select more locations than required paths and vary the threshold: We apply a higher one first and switch to the lower distance threshold after reaching the required number of paths or as soon as there are no feasible locations to add anymore.

\subsubsection{From cost mask to graph}
\label{sec: mask to graph}
We now interpret each pixel as a vertex and edges are added only between diagonally or horizontally neighboring vertices and only directed to the right. The cost of an edge is the sum of the costs of both connected pixels, where the costs of the diagonal edges are tripled afterwards to penalize detours.\newline

In mathematical terms, let $G(V,E)$ denote the derived \struc{weighted, directed graph} with vertices $V:=\{v_{ij}: i\in \{1,...,H\}, j\in \{1,...,W\}\}$ and edges $E:=\{e_{v_{ij}, v_{kl}}: l-j = 1, |k-i| \leq 1\}$.
The edge weights are given by $c:E\to \mathbb{R}_{>0}$ with \newline
\begin{align}
\notag c(e_{v_{ij},v_{kl}}) = \left\{\begin{array}{ll}C_{i,j}+C_{k,l} & \, \textrm{if}\quad  i=k \\ 3* (C_{i,j}+C_{k,l}) & \, \textrm{else} \\
\end{array}\right. 
\end{align}
Based on $G(V,E)$, Dijkstra's algorithm \cite{Dijkstra.1959} yields the cheapest paths from all start points to all points on the right border.

\subsubsection{End point and path selection}
\label{sec: end point and path selection}
\begin{figure}[h]
\captionsetup{margin={0.2cm, 0.2cm}}
  \centering
    \includegraphics[width=0.26\textwidth]{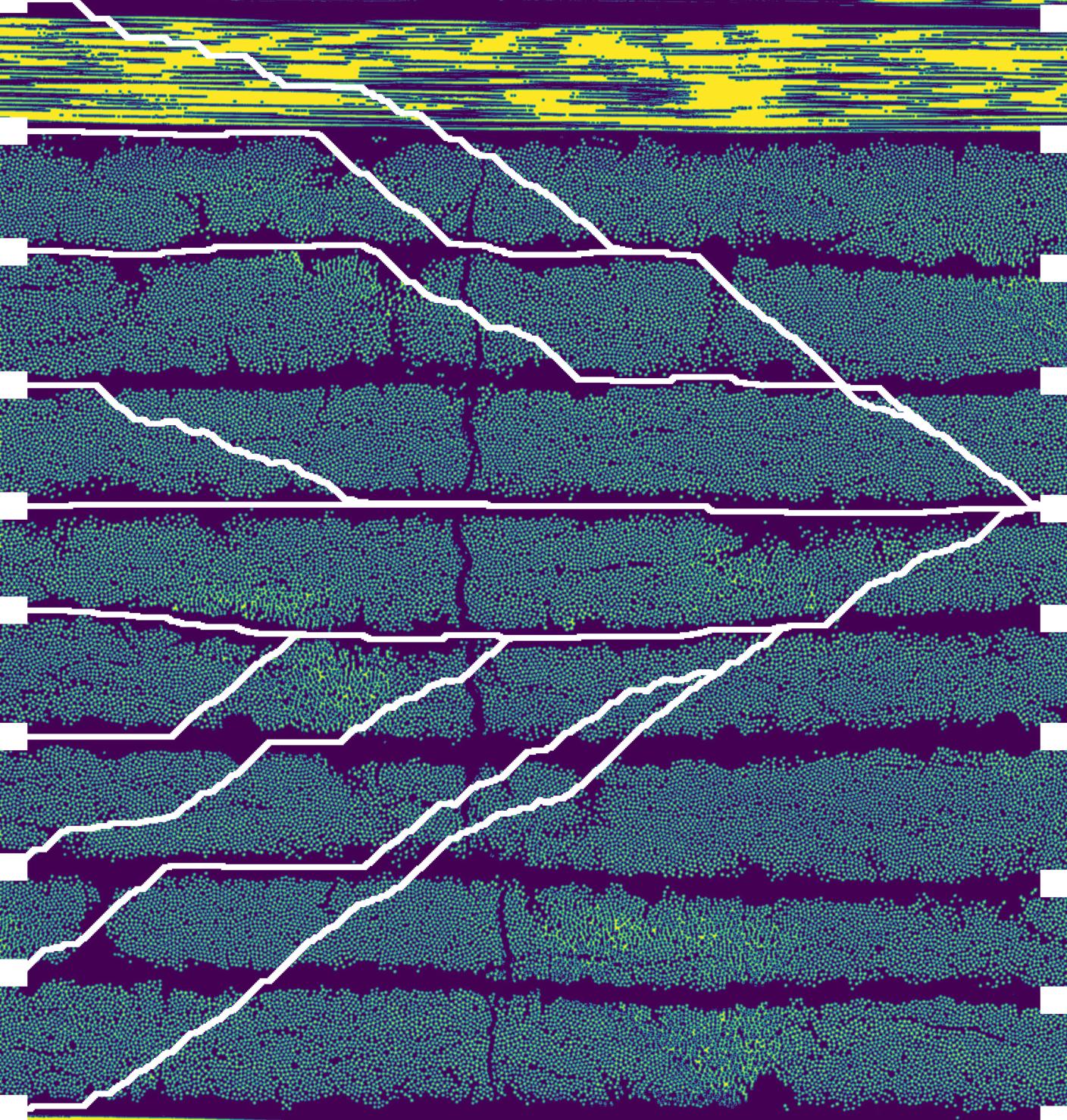}
    \caption[Cost mask with start points on the left, end points on the right, and shortest paths from one end point to all start points in white.]{Cost mask with start points on the left, end points on the right, and shortest paths from one end point to all start points in white.\footnotemark}
    \label{Fig: Process_2}
\end{figure}

Then, the end points on the right are selected the same way as the starting points, but not based on local costs, but on the cost of their cheapest path to any start point (see \cref{Fig: Process_2}). Next, we determine all feasible subsets of start and end vertices according to the stricter, i.e., higher, minimum distance threshold that contain as many elements as we need paths. For each of them, we calculate a bipartite matching for start and end points based on the path costs between them. We select the final start and end points as well as the matching according to the sum of all path costs. Last but not least, we multiply the resulting path locations by our downsampling factor of four such that they match the original micrograph and segmentation mask. \newline 

\section{Results and discussion}
\subsection{Evaluation}
\label{sec: evaluation}
\footnotetext{Two of the upper paths on the right appear to separate and later cross each other (from right to left), however, investigating this plot with maximum resolution, the paths do not meet a second time.}
We validate our approach qualitatively on the ten micrographs described in \cref{sec: data acquisition}. To distinguish correctly between ply instances, either interleaf layers or different fiber orientations in adjacent plies are required. Moreover, all plies must cross the left and right border, i.e., gaps at the border or left and right specimen edges inside the image are unfeasible. In the specimens of \cref{fig:H,fig:I,fig:J}, the two middle plies have the same orientation and are not visually distinguishable which is why we only aim to detect all remaining ply-separating paths in these cases. \newline
The following, individual parameters must be provided per segmentation mask: the number of paths to detect, the approximate minimum ply height to define minimum distance between neighboring start or end points, and optionally the rotation angle to compensate for inclinations. \newline

\begin{figure}[p]
    \centering
    \begin{subfigure}{0.32\textwidth}
        \centering
        \fbox{%
        \includegraphics[width=\linewidth,keepaspectratio]{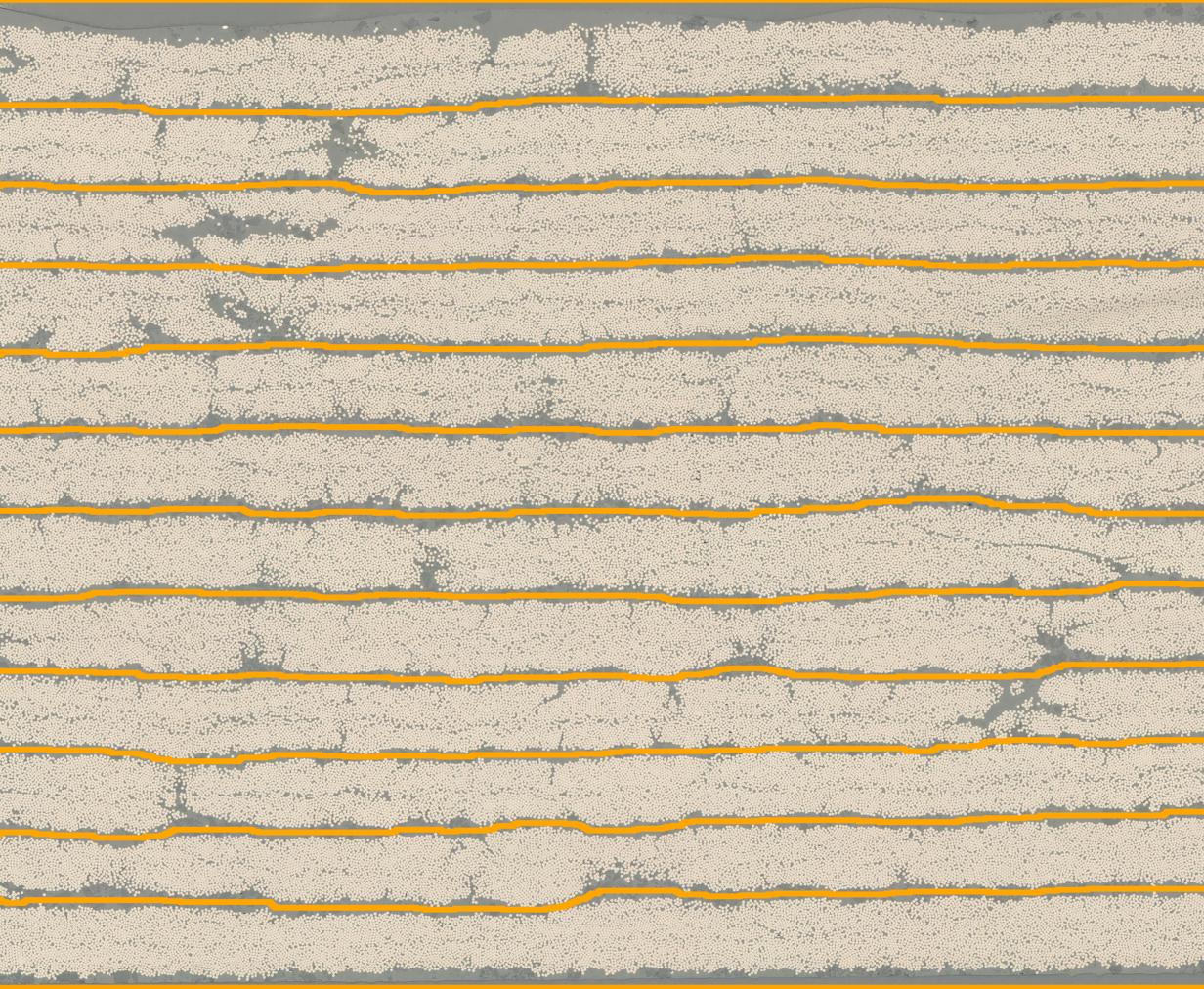}
        }
        \caption{}
        \label{fig: result_A}
    \end{subfigure}
    \hfill
    \begin{subfigure}{0.32\textwidth}
        \centering
        \fbox{%
            \includegraphics[width=\linewidth,keepaspectratio]{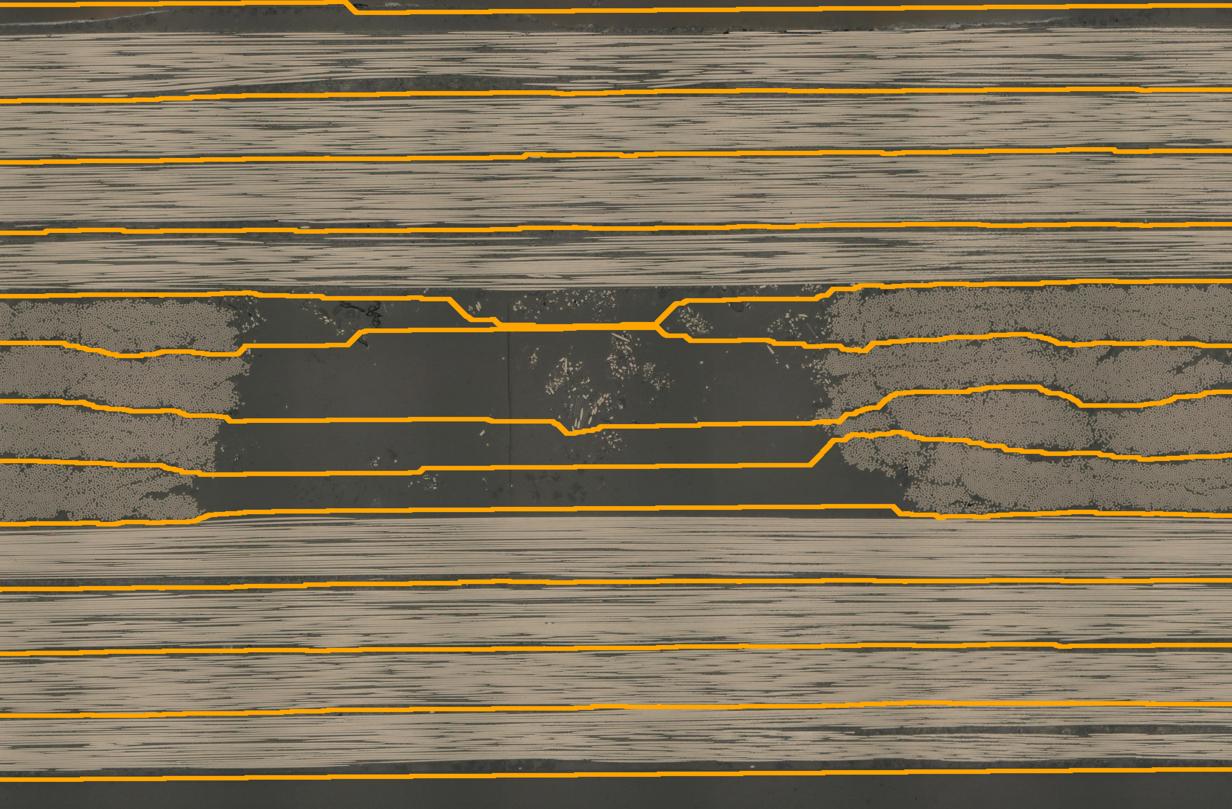}%
        }
        \caption{}
        \label{fig: result_B}
    \end{subfigure}
    \hfill
    \begin{subfigure}{0.32\textwidth}
        \centering
        \fbox{%
            \includegraphics[width=\linewidth,keepaspectratio]{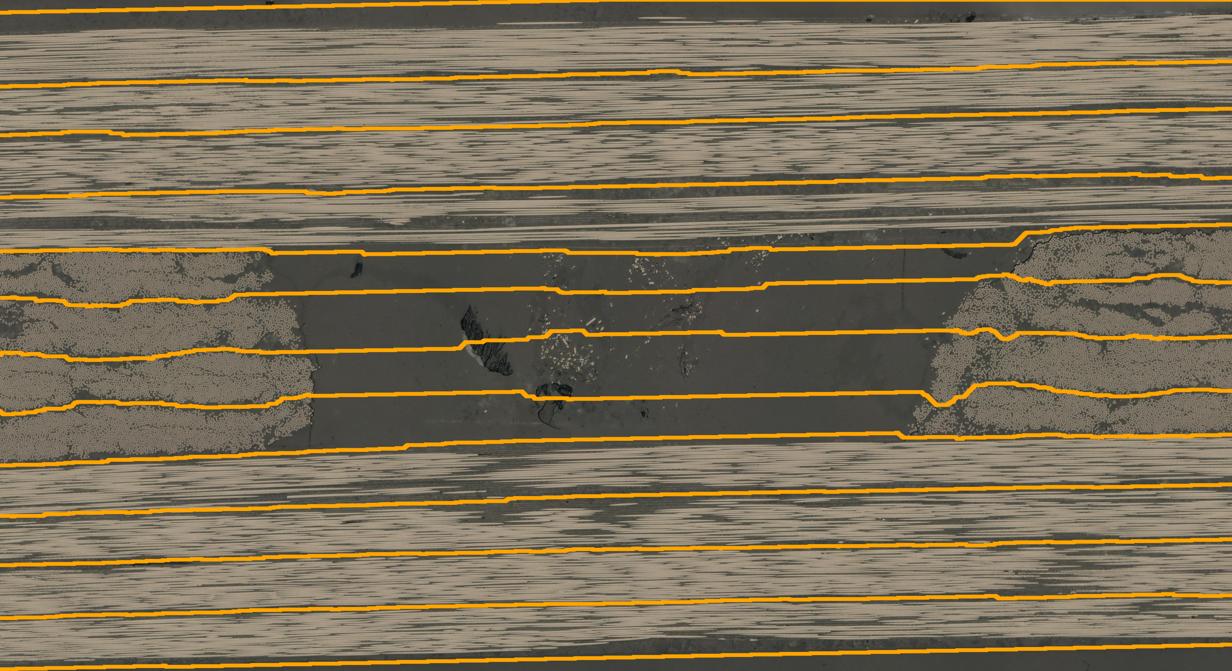}%
        }
        \caption{}
        \label{fig: result_C}
    \end{subfigure}

    \vspace{0.5cm}

    \begin{subfigure}{0.32\textwidth}
        \centering
        \fbox{%
            \includegraphics[width=\linewidth,keepaspectratio]{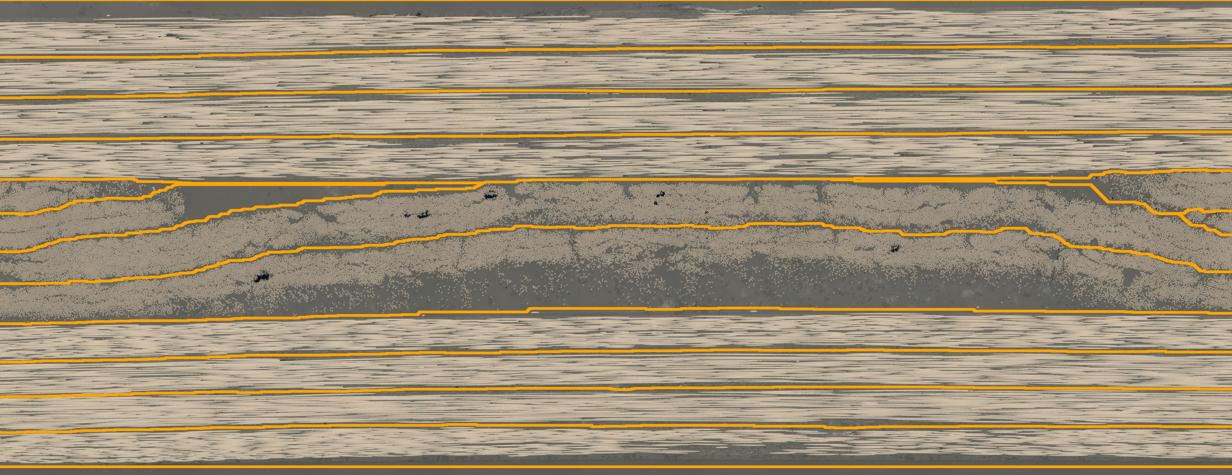}%
        }
        \caption{}
        \label{fig: result_D}
    \end{subfigure}
    \hfill
    \begin{subfigure}{0.32\textwidth}
        \centering
        \fbox{%
            \includegraphics[width=\linewidth,keepaspectratio]{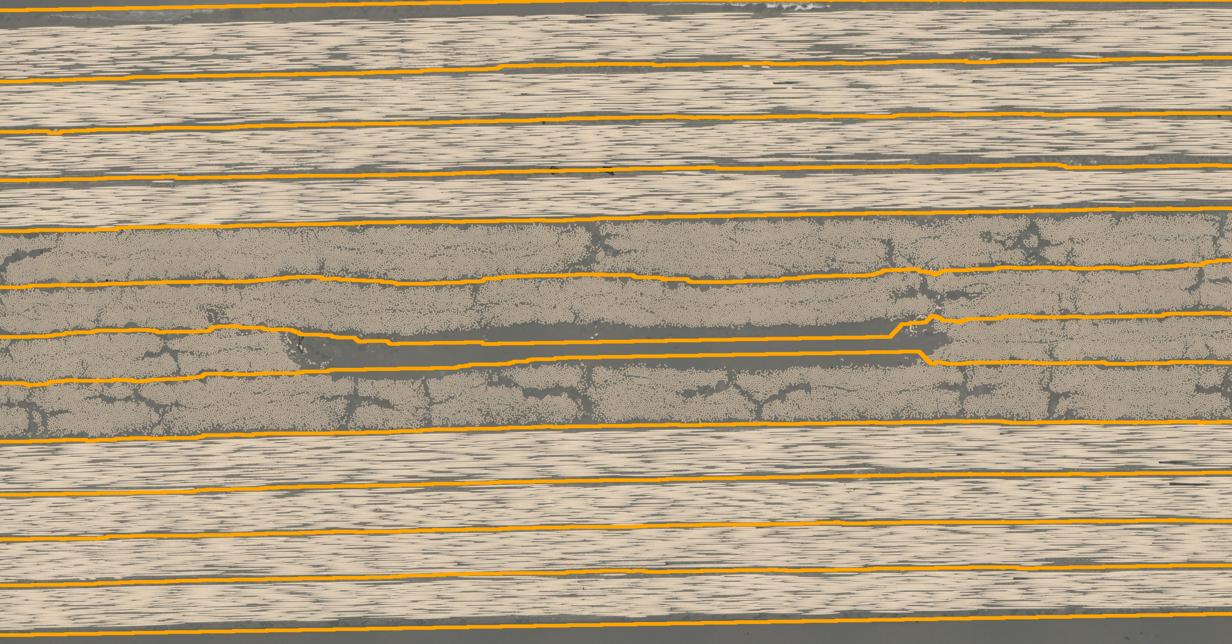}%
        }
        \caption{}
        \label{fig: result_E}
    \end{subfigure}
    \hfill
    \begin{subfigure}{0.32\textwidth}
        \centering
        \fbox{%
            \includegraphics[width=\linewidth,keepaspectratio]{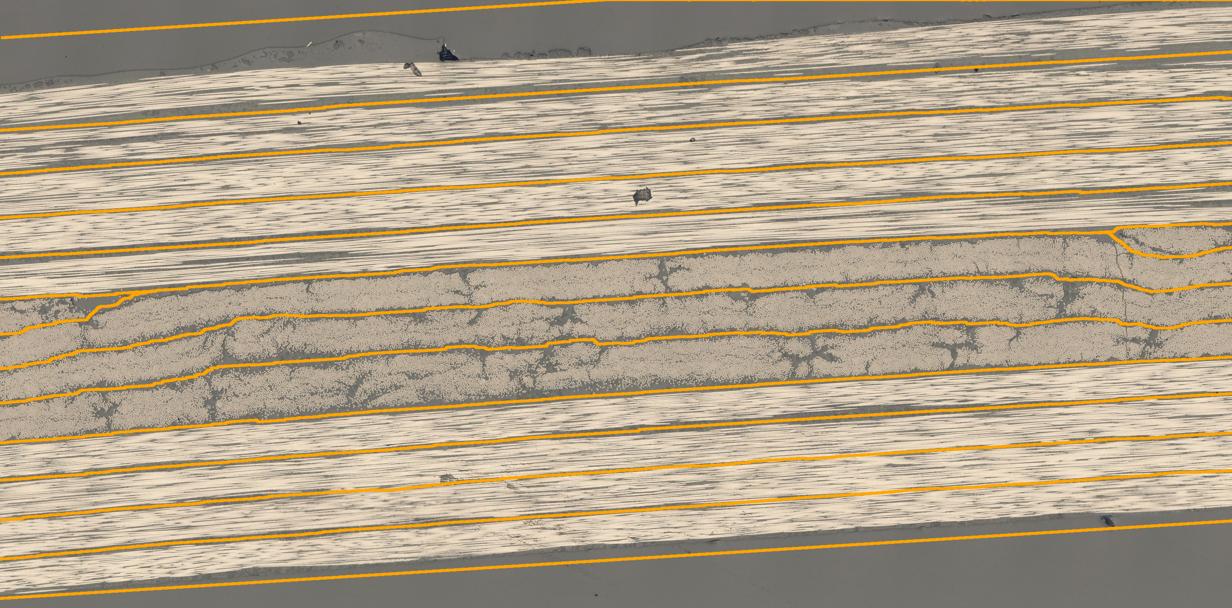}%
        }
        \caption{}
        \label{fig: result_F}
    \end{subfigure}

    \vspace{0.5cm}

    \begin{subfigure}{0.19\textwidth}
        \centering
        \fbox{%
            \includegraphics[width=\linewidth,keepaspectratio]{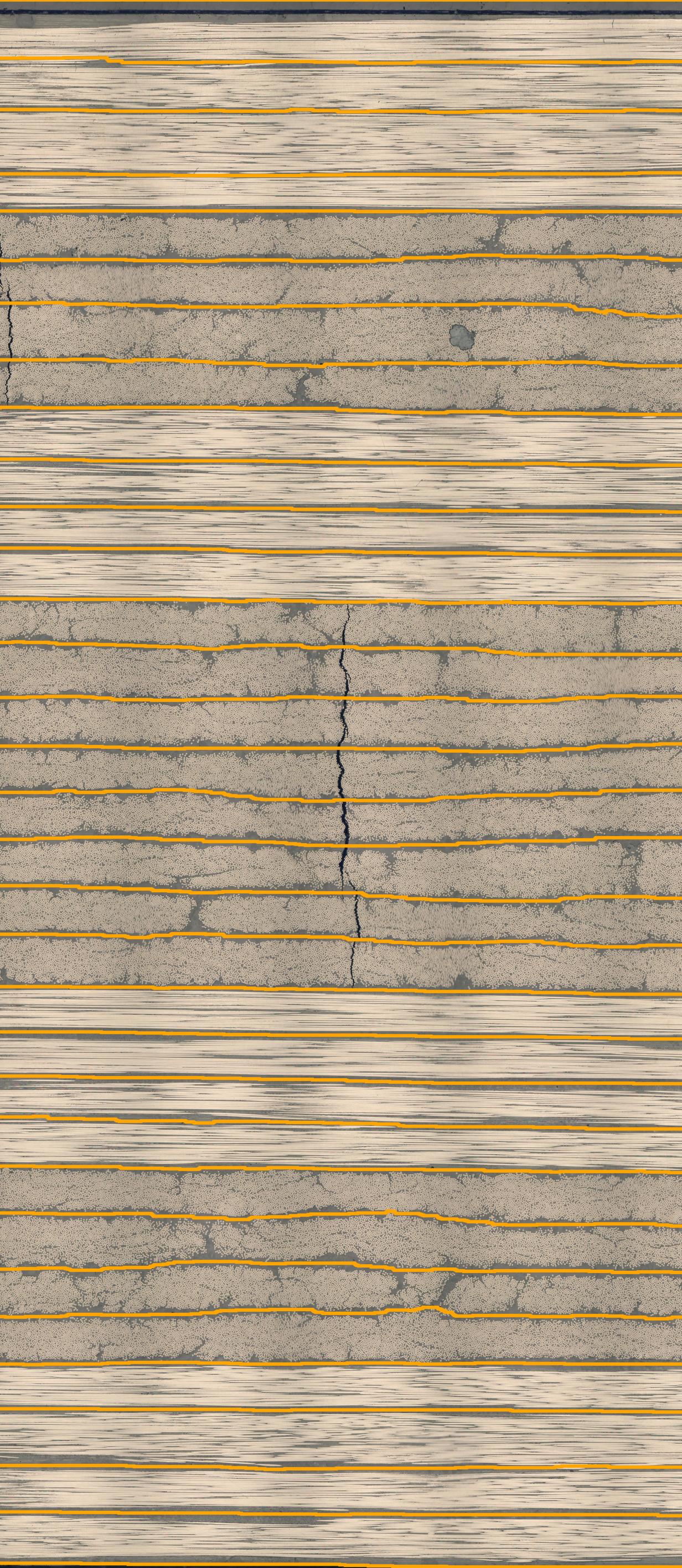}%
        }
        \caption{}
        \label{fig: result_G}
    \end{subfigure}
    \hfill
    \begin{subfigure}{0.25\textwidth}
        \centering
        \fbox{%
            \includegraphics[width=\linewidth,keepaspectratio]{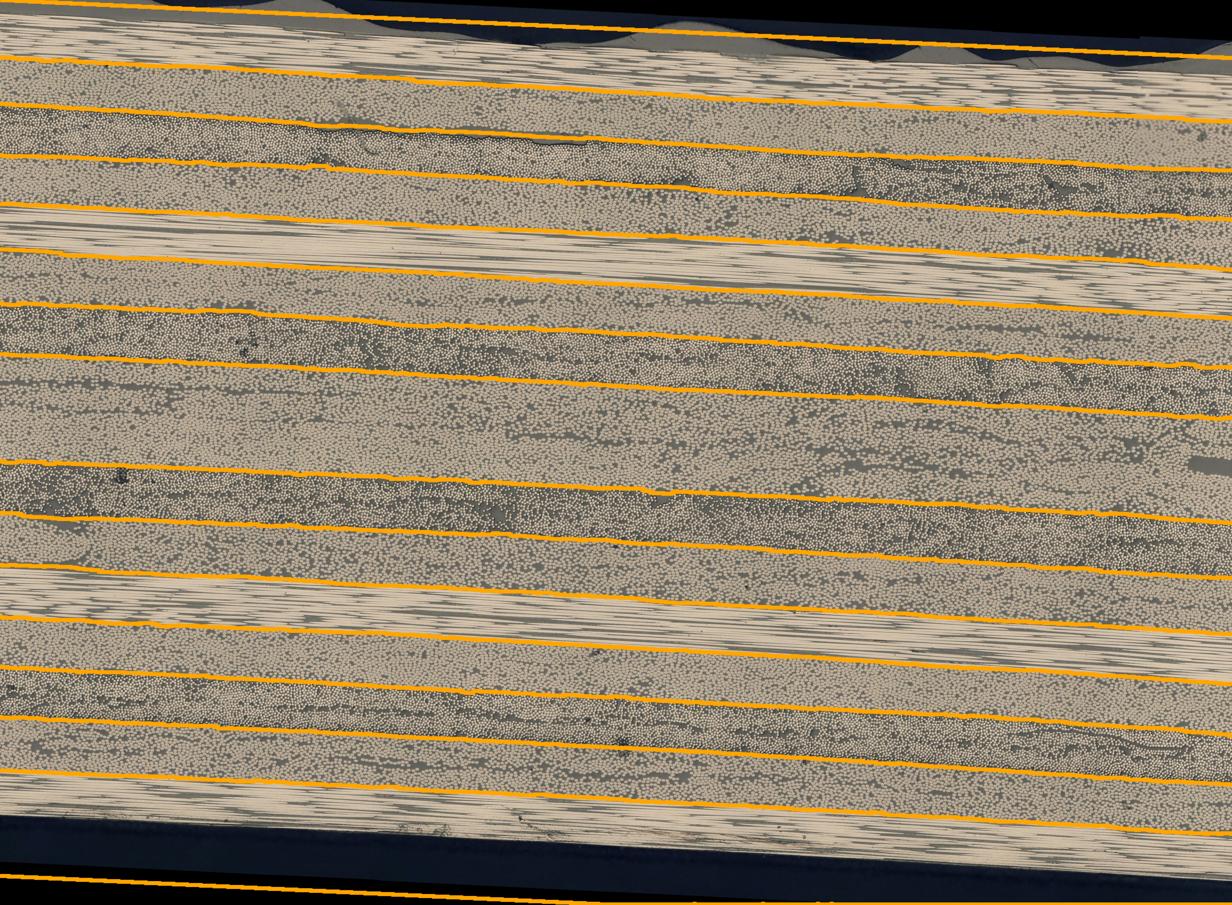}%
        }
        \caption{}
        \label{fig: result_H}
    \end{subfigure}
    \hfill
    \begin{subfigure}{0.25\textwidth}
        \centering
        \fbox{%
            \includegraphics[width=\linewidth,keepaspectratio]{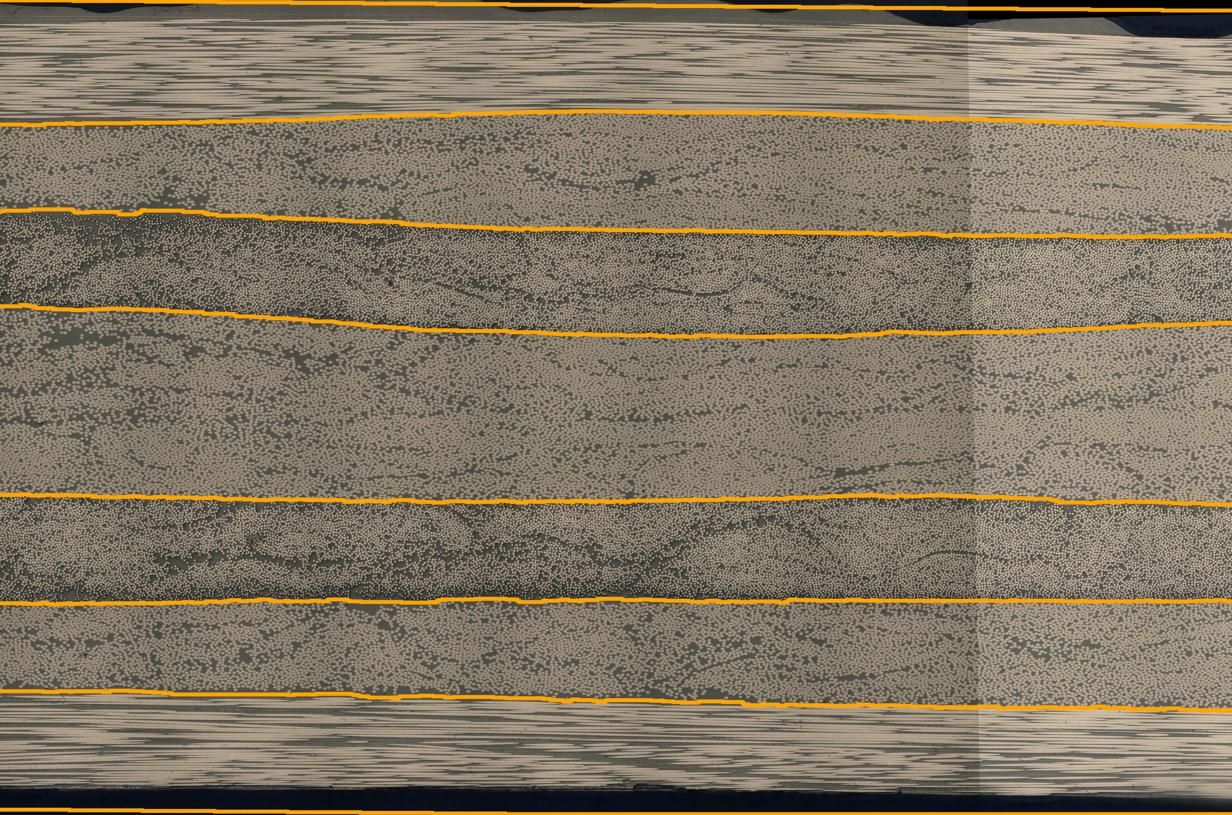}%
        }
        \caption{}
        \label{fig: result_I}
    \end{subfigure}
    \hfill
    \begin{subfigure}{0.25\textwidth}
        \centering
        \fbox{%
            \includegraphics[width=\linewidth,keepaspectratio]{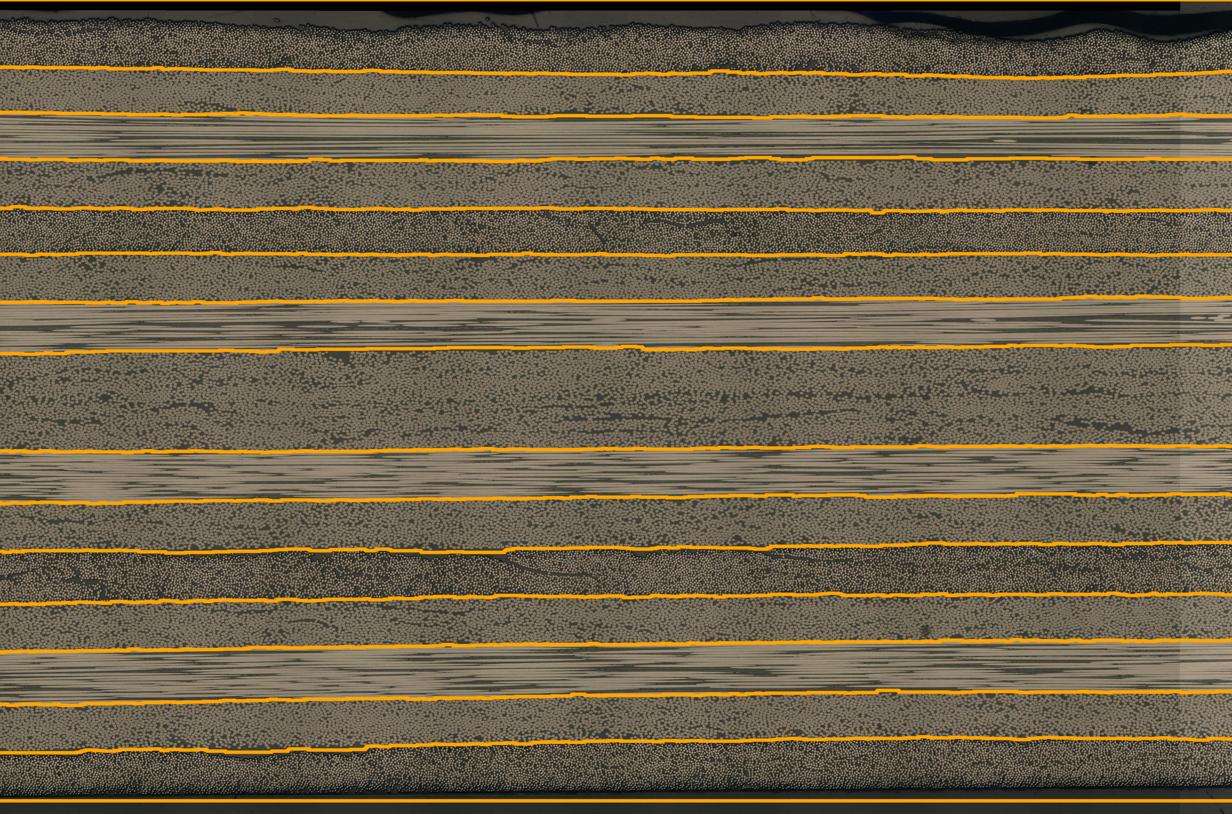}%
        }
        \caption{}
        \label{fig: result_J}
    \end{subfigure}

    \caption{The evaluation results based on the segmentation mask obtained by our segmentation model.}
    \label{fig:overview_results}
\end{figure}

Our approach determines the ply-separating paths in all micrographs correctly. Here, broken fiber pieces in gaps are not counted as errors, as their assignment is not possible.
On these ten examples, the application takes between 9 and 38 seconds\footnote{with a Python 3.14.3 setup in Visual Studio Code on an Intel\textsuperscript{\textregistered} Core\textsuperscript{\texttrademark} i7-1370P processor with 32 GB RAM.} (avg. 21.58 s) excluding the downstream analyses. As a function of the segmentation mask size, the approach takes 0.41 to 0.64 seconds per one million pixels. The results are visualized in \cref{fig:overview_results}.\newline

\afterpage{\clearpage}

Using the binary segmentation masks obtained by Otsu's method yields almost identical, flawless results for the seven specimens with interleaf layers. However, Otsu's method is not well suited to distinguish between different fiber orientations and therefore the usage of the corresponding segmentation masks here is restricted to specimens with interleaf layers. Here, the application takes between 8 and 35 seconds (avg. 19.57 s) again excluding the downstream analyses. As a function of the segmentation mask size, the approach takes 0.38 to 0.59 seconds per one million pixels. The minimal time gain on average is expected to be caused by fact that the EDT does not need to be calculated separately for different fiber classes. \newline

\subsection{Challenges and approaches to solutions}
\label{sec: challenges}
This section provides insights from the development process, where incorrect paths occurred, to show limitations and describe adjustments that solved the respective issue in our case or can be helpful for future data. \newline

Our approach can yield \textit{ply-crossing detours}, i.e., a path crosses a ply twice to benefit from a cheaper region in between, which can pay off especially for a high ratio of micrograph width and ply thickness.
There are different ways to make this less likely: The cost term weight $w_1$ can be increased and/or edges exceeding a set cost limit can be excluded to make crossing plies more difficult. Only using horizontal and diagonal edges to the right and omitting vertical ones is also very effective in preventing the occurrence of this problem.

\begin{wrapfigure}[15]{r}{0.3\textwidth}
\captionsetup{margin={0.1cm, 0.1cm}}
  \centering
    \includegraphics[width=0.28\textwidth]{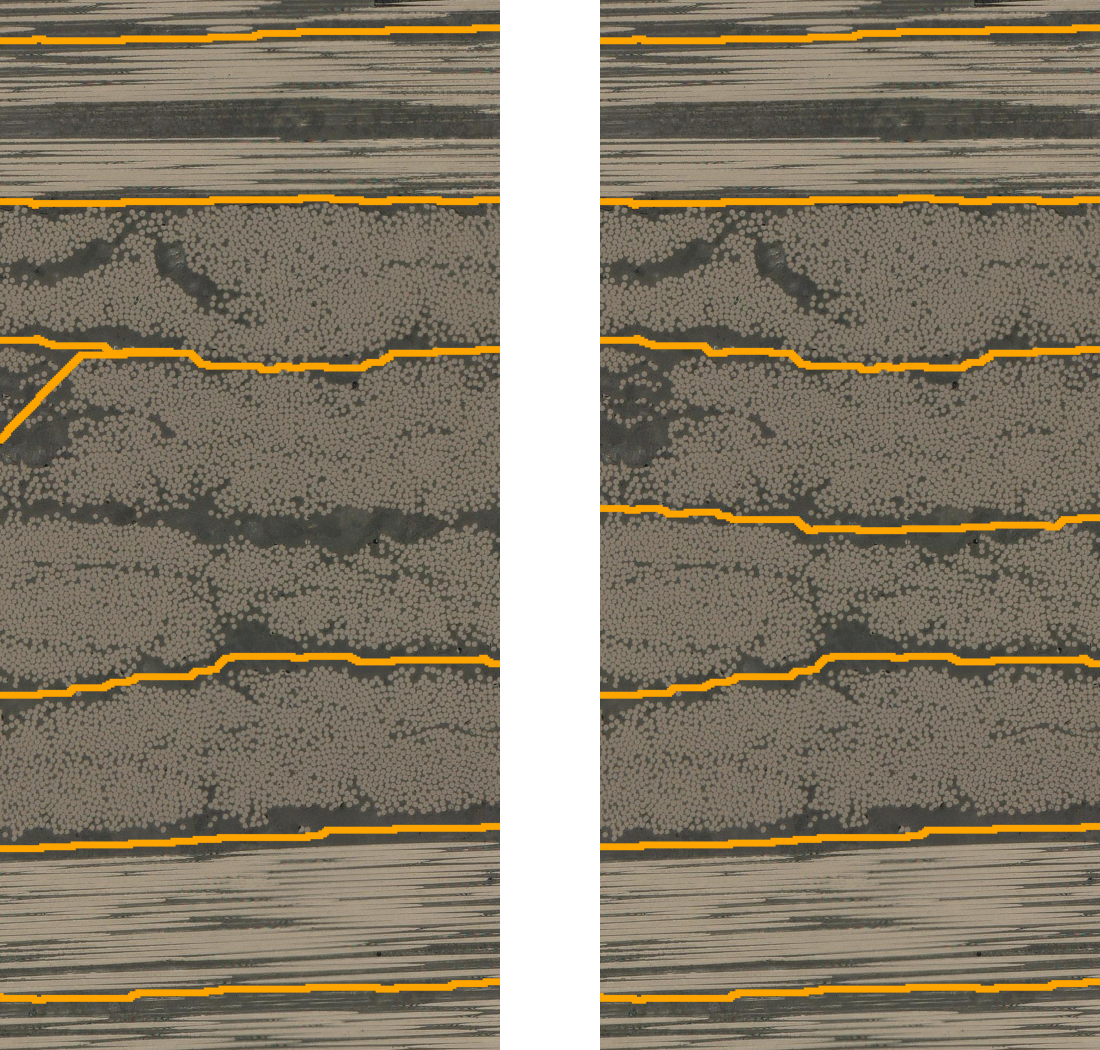}
    \caption{Fixing an incorrect start location that led to an incorrect path.}
    \label{Fig: challenge 2}
\end{wrapfigure}
As explained in \cref{sec: start point calc}, mistakes during the start point selection can occur since the decision is made based on local information (see \cref{Fig: challenge 2}, left). Selecting more start points than needed and finally sort out the surplus ones based on the path costs including global information has solved this problem mostly. If it still occurs, potential solutions include a further increase in the number of initial start points or an adjustment of the border column width that controls the start point calculation's extent of locality (see \cref{Fig: challenge 2}, right). \newline
Another challenge are \textit{unfavorable resin ramifications} representing a cheap path inside a ply. Setting a higher approximate minimum ply thickness, i.e., a stricter minimum distance threshold for the start and end points can help, but is not always suitable. \newline
The reliability of the approach can be further increased by specifying the start and end points or, if user interactions are permitted, by correcting them after automatic calculation if required.\newline

There are even more parameters that have been kept constant here, but can easily be adjusted if required for future data. Firstly, the downsampling factor that can be adjusted especially for different pixel sizes to find a good trade-off between sufficiently high resolution and efficiency. 
Moreover, the minimum FVF in the column slice between different start points can be increased to prevent two start points in a resin-rich area between the same plies from being both considered. 
Using a directed graph here means that all edges point to the right side enforcing a rather direct path than detours while undirected graphs can be required for more advanced path requirements. 
Last but not least, the weights for the different cost terms can be customized to the respective data. The third cost term $C_3$, weighted by $w_3$, is only used for the path calculation in case of no interleaf where neighboring plies must have a different fiber orientation to be distinguishable. If we would set $w_3=1$ for \cref{fig:B,fig:C,fig:D,fig:E,fig:F,fig:G}, the paths between plies of same fiber orientation could be diverted to the two interleaves between different fiber orientation. For the start point calculation, there is no such risk and therefore we always use the information about different nearby fiber orientations, i.e., $C_3$, if those exist.\newline
An overview of these parameters and the corresponding values is provided in \cref{tab: adjustable parameters}.
A thorough analysis or optimization of these parameter values is out of scope of this paper. 

\begin{center}
\begin{table}
\caption{Overview of parameters to provide per input (\textit{micrograph-dependent}) and to adjust as required. The latter are kept constant for all segmentation masks during evaluation.}
\label{tab: adjustable parameters}
\centering
 \begin{tabular}{ cc } 
\toprule
 Parameter & Value \\ 
 \midrule
 Number of plies & \textit{micrograph-dependent} \\ 
 Approximate minimum ply height & \textit{micrograph-dependent} \\
 Rotation angle & \textit{micrograph-dependent} \\ \hline
 Fiber radius $r_{\text{fiber}}$ & 6\,px \\
 Approx. average interleaf layer thickness $t_{\text{interleaf}}$ if existing & 60\,px \\
 Downsampling factor & 4 \\
 Border column width for start point calculation & 400 px \\
 Maximum feasible edge cost & None \\
 Directed or undirected graph & directed \\
 Minimum FVF between start/end points & 0.1 \\
 Upscaling factor for number of start/end points & 1.5 (limited to 5 additional) \\ \addlinespace
 Cost term weight $w_1$ & $
            \begin{cases} 
                1&\text{if no interleaf}\\
                100&\text{else} \\
            \end{cases}
            $
\\ \addlinespace 
 Cost term weight $w_2$ & 1 \\ \addlinespace
 Cost term weight $w_3$ & $
            \begin{cases} 
                1&\text{if multiple fiber classes}\\
                0&\text{else} \\
            \end{cases}
            $
\\ \addlinespace
 Cost term weight $w_3$ for start point calculation & 1 \\
 \bottomrule
\end{tabular}
\end{table}
\end{center}

\section{Downstream ply analyses}
\label{sec: downstream analyses}

\begin{figure}
\begin{subfigure}[h]{0.8\linewidth}
\centering
    \includegraphics[width=\textwidth]{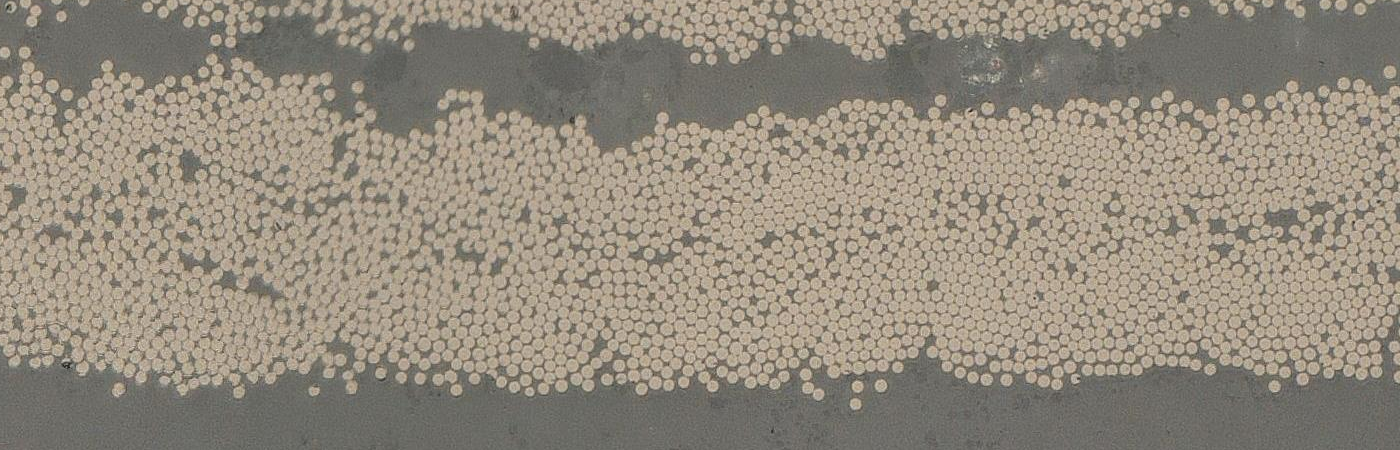}
\end{subfigure}
\begin{subfigure}[h]{0.8\linewidth}
\centering
    \includegraphics[width=\textwidth]{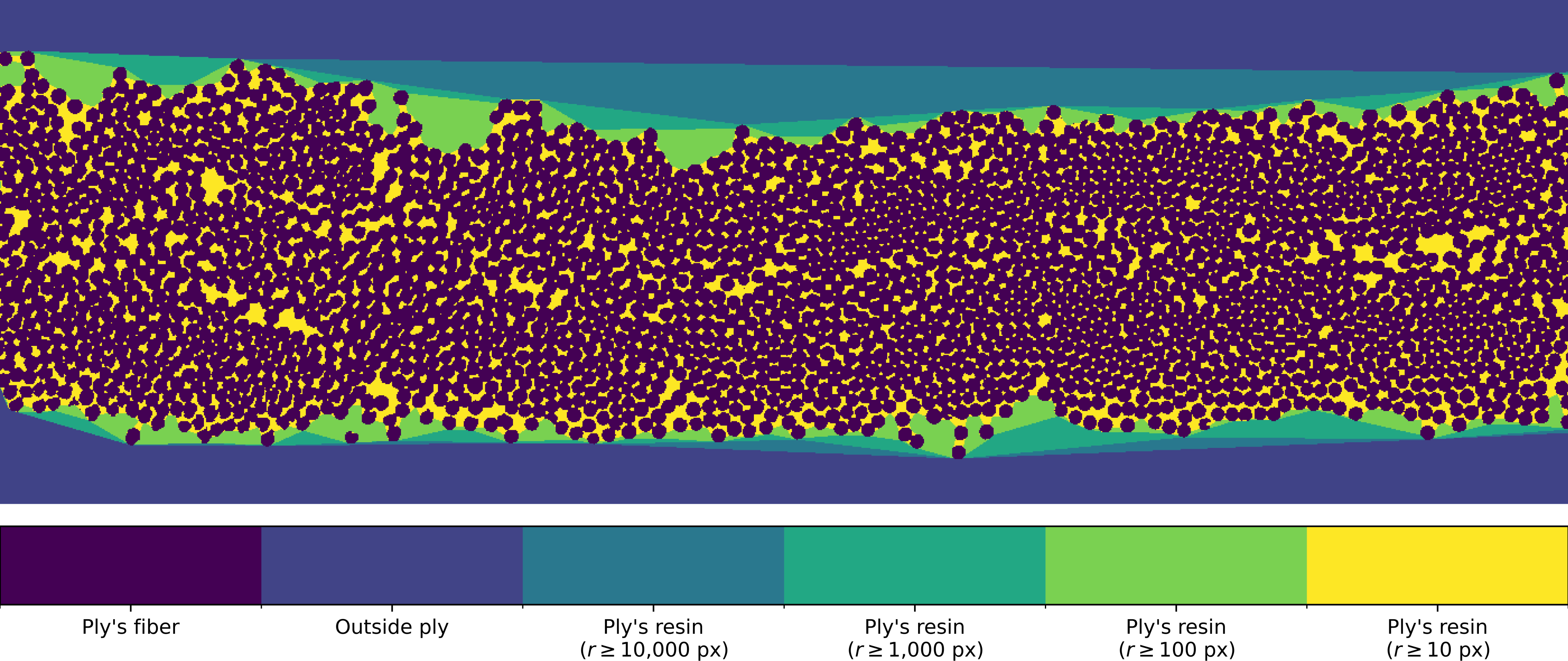}
\end{subfigure}
\caption{Visualization of the effect of different $r$- and, consequently, different $\alpha$-values on the interior face of the corresponding $\alpha$-shape given a ply's fiber pixels as point cloud.}
\label{fig: alphashape}
\end{figure}

All of the following use cases have in common that they require the determination of layer boundaries. While ply-separating paths allow fibers to be assigned to specific plies, they are not necessarily suitable for defining the boundary of a layer. Instead, an $\alpha$-shape \cite{Edelsbrunner.1983} is employed which represents a generalization of the convex hull. The $\alpha$-shape (defined below in \cref{def:alphashape} for $\alpha < 0$) encloses the set of a ply's fiber locations (either all fiber pixels or fiber centroids) with a parameter $\alpha$ controlling the tightness of the boundary. A visualization of different $\alpha$-values can be found in \cref{fig: alphashape}. As fiber locations, we can either use all fiber pixels or - in case of circular 0° fibers - only the fiber centroids determined as local maxima in $EDT_{\text{resin}}(M)$. In the latter case, we need to add a buffer afterwards to shift the $\alpha$-shape outwards by the fiber radius. $\alpha$ should be selected depending on the number of pixels per fiber diameter and the FVF. The values used here are only examples chosen to indicate the potential for the respective use cases. \newline
To measure heights of plies and interleaf layers in what follows, we require one post-processing step after calculating the $\alpha$-shape: For every horizontal location, we calculate the minimum and maximum vertical pixel location inside the $\alpha$-shape as lower and upper boundary. Thereby, we prevent interruptions in the perpendicular distance measurements that are possible in the $\alpha$-shape itself, i.e., for a given x-position, between the lower and upper boundary, there can be pixels not belonging to the area inside the $\alpha$-shape.\newline 

\begin{definition}[\cite{Edelsbrunner.1983}] 
	Let $S$ be a set of points in the plane and $\alpha < 0$. The \struc{$\alpha$-shape} of $S$ is a (straight line) graph $G(V,E)$ with:
    \begin{itemize}
        \item $V \subseteq S$ containing all points $p$ of $S$ for which there exists a disc of radius $r=\frac{1}{|\alpha|}$ with $p$ lying on its boundary such that the disc's closed complement contains all points of $S$.
        \item $E$ contains all edges between vertices $p,q \in V$ for which there exists a disc of radius $r=\frac{1}{|\alpha|}$ with $p,q$ lying both on its boundary such that the disc's closed complement contains all points of $S$.
    \end{itemize}
    \label{def:alphashape}
\end{definition}

\subsection{Fiber volume fraction distribution}
\label{sec: fvf}

The FVF is a key structural parameter, as it defines the relative proportion of load-bearing fibers with respect to the matrix and thereby influences local elastic properties as well as strength-related performance. Conventional methods for FVF determination, such as matrix removal (burn-off) or density-based approaches \cite{ASTM_D3171_2022, ISO_14127_2024}, generally provide integral values for an entire specimen volume. While these methods are indispensable as quality indicators, they do not capture the distribution of FVF across individual plies.

For many applications, however, this local distribution is of particular importance. Previous studies based on micrograph evaluations have shown that real laminate and tape microstructures can deviate substantially from an ideally uniform fiber distribution \cite{Appels.2025}.

A ply-wise analysis of the FVF distribution requires reliable identification of the individual ply boundaries. Based on the aforementioned $\alpha$-shape approach, a segmentation mask with four classes is obtained: outside ply, resin, fiber, and other (e.g., cracks or unidentified regions). Subsequently, a heatmap stating the local FVF for a neighborhood around each pixel is calculated following \cite{Naumann.2025} and a histogram of the FVF distribution is generated as described in \cite{Schafer.2024}. Exemplary resulting plots are shown in \cref{fig:Process} (bottom).

The resulting ply-resolved FVF distributions enable advanced analyses by capturing local variations in fiber content that remain hidden in laminate-average values. As a result, the elastic constants $E_1$, $E_2$, and $G_{12}$ can be estimated more accurately and individually, since they depend directly on the local fiber volume fraction. In addition, the estimation of matrix-dominated properties and local stress states, including transverse strength and interlaminar stresses, becomes more reliable because these quantities are highly sensitive to local FVF variations and heterogeneous fiber packing \cite{Li.2024,Hojo.2009,Cater.2018}. This is particularly relevant for CFRP laminates used in cryogenic applications, where microcrack initiation is strongly influenced by residual stresses generated during cool-down, as fibers, matrix and differently oriented plies exhibit different thermal contractions while being mechanically constrained from deforming freely. Ply-resolved FVF maps therefore enable the identification of plies that are particularly susceptible to transverse cracking under thermal cycling, as well as the investigation of crack density as a function of ply-level FVF.

From a manufacturing and digitization perspective, automated ply-level FVF determination provides reproducible quality metrics that extend beyond laminate-average descriptions, thereby supporting process control and digital material characterization workflows.

\subsection{Measurement of interleaf resin layer thicknesses}
\label{sec: interleaf}

Interleaving concepts are widely employed in modern third and fourth generation CFRP aerospace prepreg systems to improve interlaminar damage tolerance \cite{Altstaedt.1993,Shivakumar.2013,Lengsfeld.2021}. In these materials, a thin resin-rich layer - typically consisting of an epoxy matrix modified with thermoplastic particles - is located at the ply interfaces of the laminate \cite{Masters.1991,Shivakumar.2013,Lengsfeld.2021}. The presence of such interleaf layers has been shown to increase interlaminar fracture toughness and to reduce delamination growth under static and dynamic loading conditions \cite{Shivakumar.2013,Hojo.2006,Saghafi.2017}.

Reported nominal interleaf thicknesses in thermoplastic particle-modified aerospace prepregs range from approximately 25-30 µm (nominal values) \cite{Tilbrook.2011} to 30-70 µm based on microsection analyses \cite{Frerich.2019,Kappel.2020}. Micrographic observations indicate that interleaf thickness is not strictly uniform but exhibits local variations. Thus, in addition to the nominal thickness ranges reported for specific prepreg systems, the interleaf thickness should be considered as a spatially varying quantity within the laminate cross-section.

Since interlaminar fracture is governed by micromechanical processes acting within the resin-rich interface region - including particle debonding, matrix yielding and crack deflection \cite{Shivakumar.2013} - the local morphology of the interleaf layer influences interlaminar crack initiation and propagation. Previous studies assessing the influence of interleaf layers on fracture toughness predominantly investigated global variations of the interlayer thickness \cite{Gibson.2001,Chai.1992}. Quantitative data and analysis of spatially resolved interleaf thickness distributions within laminates are to the authors' knowledge not publicly documented.

\begin{figure}[t]
\begin{subfigure}[h]{0.58\linewidth}
\centering
\includegraphics[width=\linewidth]{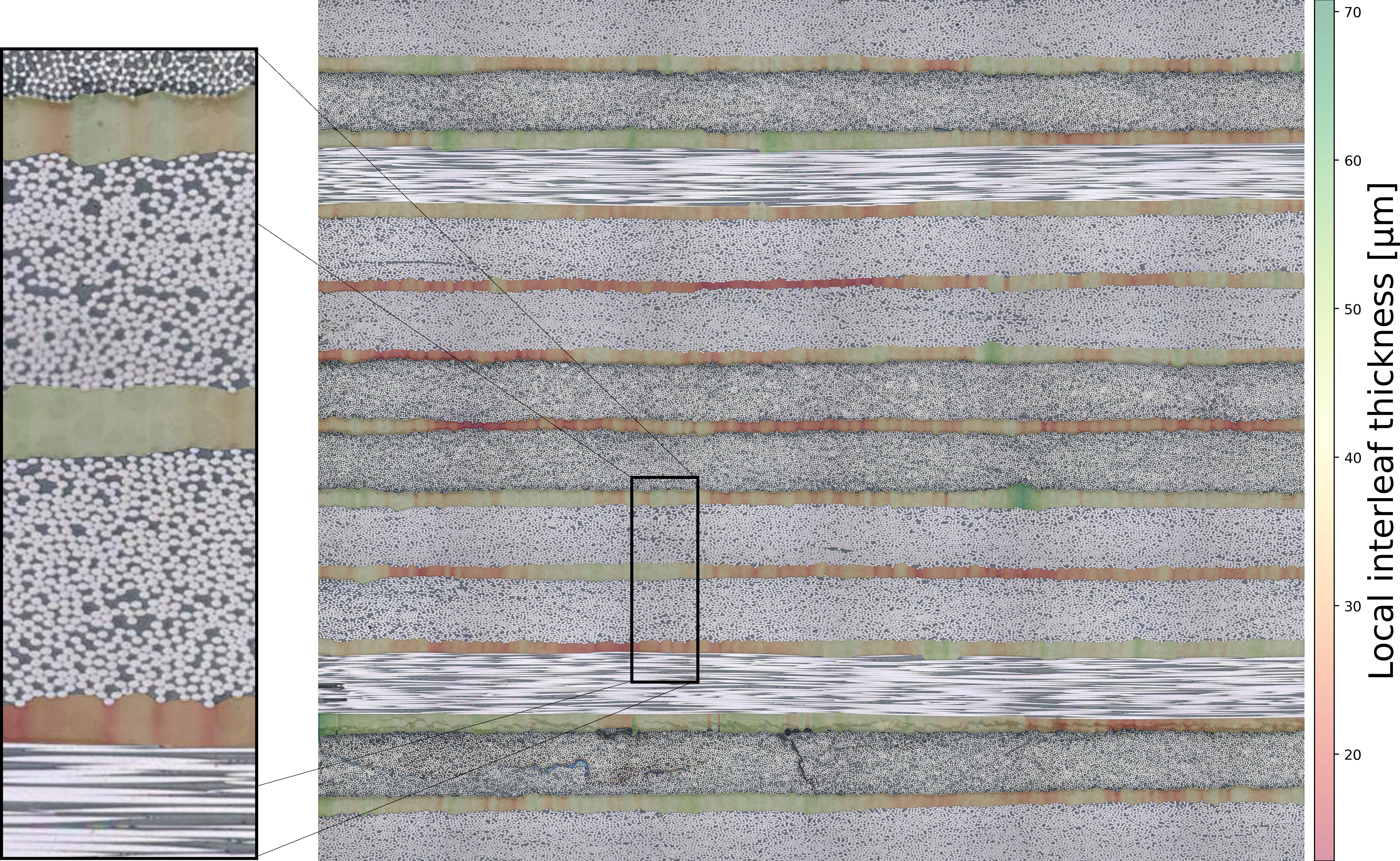}
\caption{Heatmap of local interleaf thicknesses with $r$=45 px for the $\alpha$-shape calculation.}
\label{fig: interleaf thickness heatmap}
\end{subfigure}
\hfill
\begin{subfigure}[h]{0.38\linewidth}
\centering
\includegraphics[width=\linewidth]{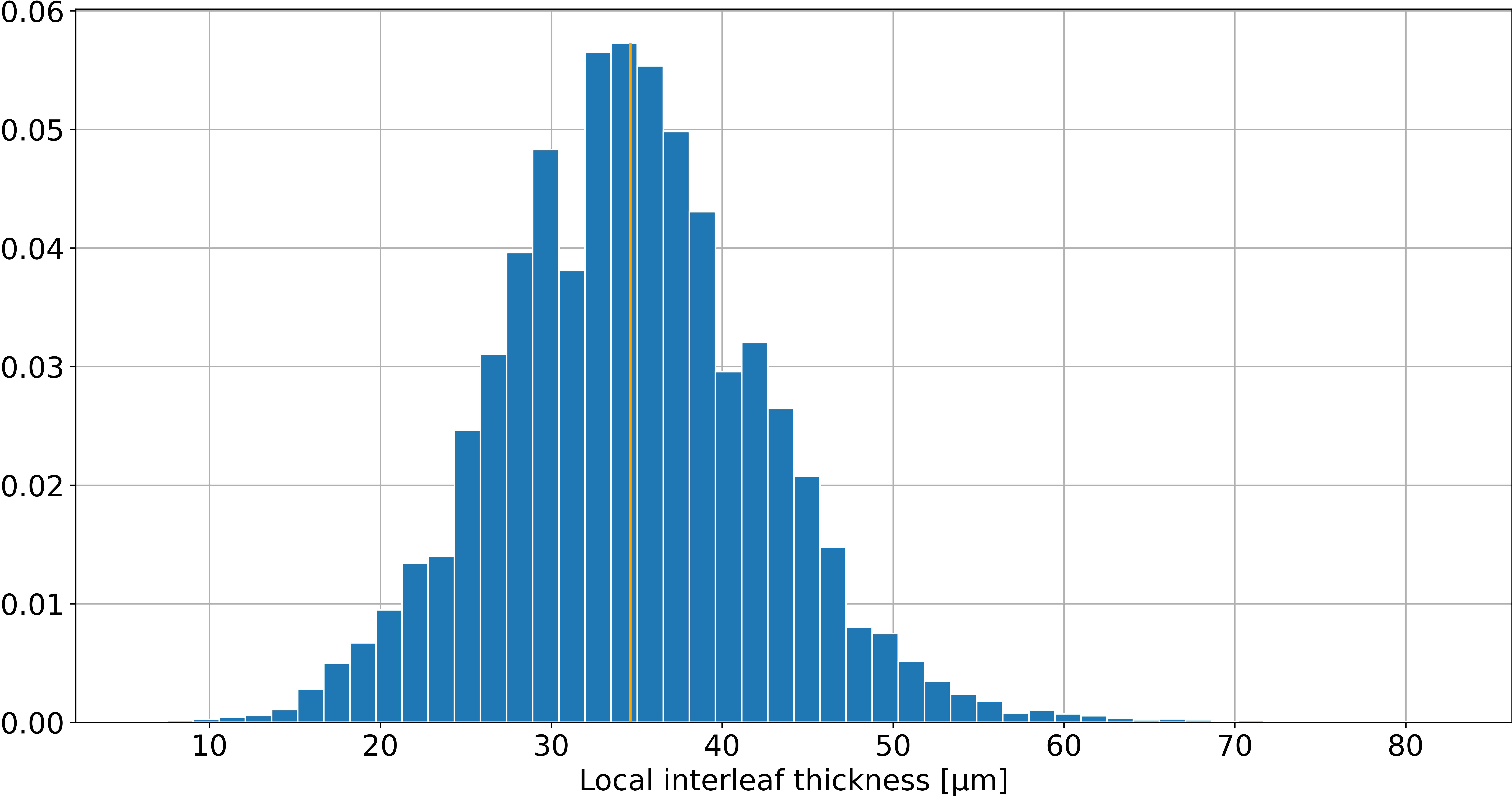}
\caption{Histogram of local interleaf thicknesses.}
\label{fig: interleaf thickness histogram}
\end{subfigure}
\caption{Visualizations of the interleaf resin layer thickness measurement corresponding to the same micrograph, but the heatmap shows only a section of it, while the histogram corresponds to the whole image.}
\label{fig: interleaf thickness analysis}
\end{figure}

The segmentation-based image analysis applied in this work enables the identification of ply interfaces and the geometric determination of local interleaf thicknesses in CFRP micrographs. After segmentation and interface tracing, perpendicular distance measurements between adjacent plies yield local thickness values (see \cref{fig: interleaf thickness heatmap,fig: interleaf thickness histogram}). By automating these steps, the approach offers a practical means to extract the as-built interleaf thickness distribution efficiently and reproducibly from larger laminate sections. The resulting data captures the spatial variability of the interleaf and provides a quantitative basis for a detailed analysis of delamination behavior.

\subsection{Measurement of fiber ply thicknesses in multidirectional laminates}
\label{sec: ply thickness}

Transverse cracks in multidirectional laminates represent a critical factor in the structural design of cryogenic hydrogen storage systems \cite{Roy.2004, Ryan.2006, Hamori.2019}. This damage mode can be characterized by the deterministic in-situ effect, which dictates that the matrix-dominated strength of a constrained ply within a multidirectional layup exceeds that of an isolated unidirectional laminate \cite{Camanho.2006, Flaggs.1982}. The magnitude of this effect is highly sensitive to the ply thickness, exhibiting a proportionality to the inverse square root of the thickness, thereby highlighting the structural advantages of thin-ply architectures \cite{Sebaey.2014, Arteiro.2018}. 

However, various manufacturing-related factors lead to local deviations in ply thickness. Particularly in thin-ply configurations, these deviations can result in significant variance in the mechanical properties critical for structural integrity \cite{Cheng.2021, Davila.2017}.

\begin{figure}[t]
    \centering
    \includegraphics[width=0.7\textwidth]{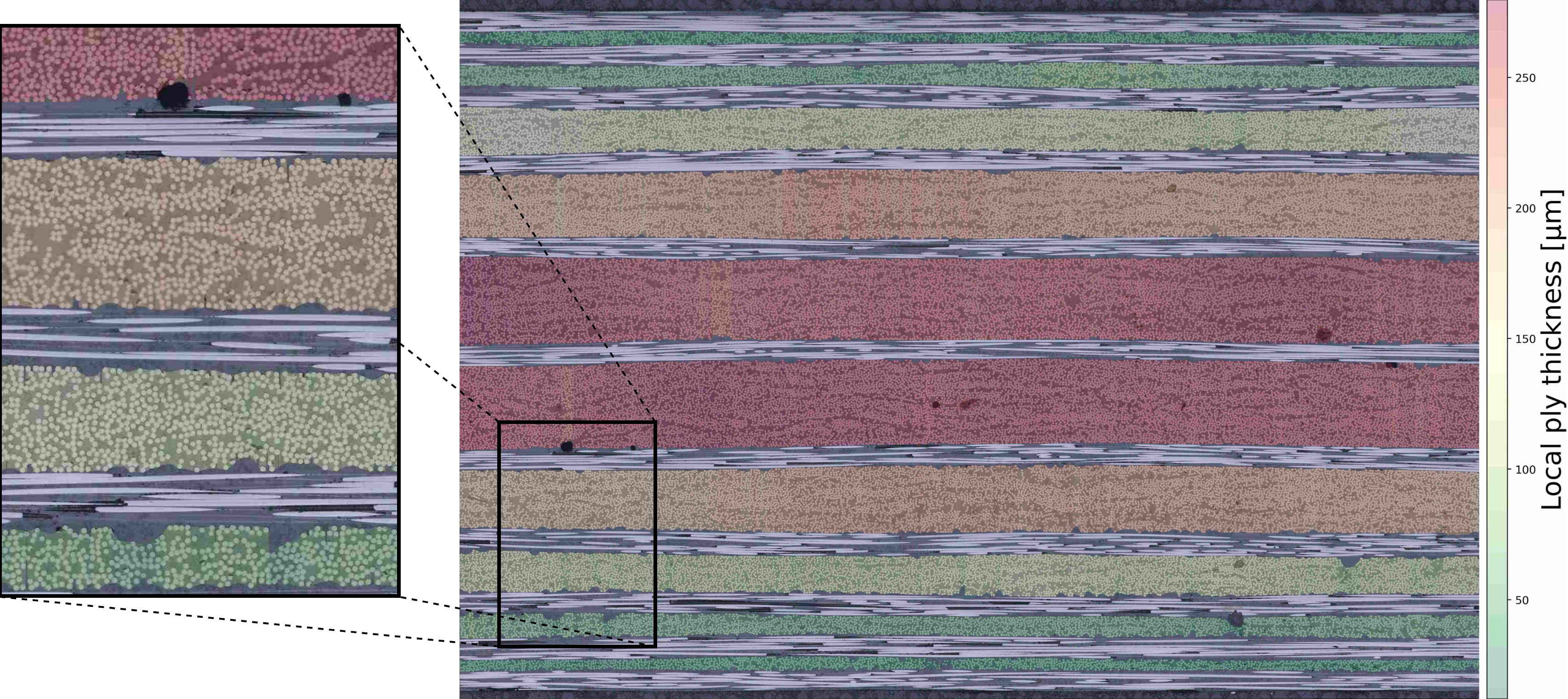}
    \caption{Heatmap of local ply thicknesses with $r$=45 px (approx. three fiber diameters) for the $\alpha$-shape calculation.}
    \label{fig: fiber ply thickness}
\end{figure}

Utilizing automated ply detection, the individual ply thicknesses can be quantified (see visualization in \cref{fig: fiber ply thickness}) within relevant test structures, allowing for the determination of deviations from the nominal design thickness. Furthermore, by subjecting material samples to iterative strain levels, this algorithmic approach - integrated with automated crack detection - enables the localized analytical determination of in-situ strengths. In the future, this methodology offers significant potential for the refinement and validation of advanced numerical models.

\section{Conclusion}
\label{sec: conclusion}
In this paper, we present an approach to automatically extract ply-specific information from segmentation masks of CFRP micrographs via ply-separating paths extracted with Dijkstra's algorithm. 
Our method enables a ply-specific analysis of the fiber distribution and automated measurements of fiber ply or interleaf layer thicknesses for which we both provided use cases in current research.\newline 
Using segmentation masks yielded by our machine learning-based model, on all ten selected micrographs covering a broad range of characteristics, the approach detects all paths correctly. For binary segmentation masks originating from Otsu's method, the applicability is limited to specimens with interleaf layers, for which the results are just as well. For other, potentially challenging micrographs, there exist a lot of adjustable parameters for users which we have roughly outlined. \newline
Future work aims for a broader applicability and fewer requirements for the characteristics of the specimens and micrographs. Moreover, comparisons with alternative approaches like panoptic segmentation or clustering based on fiber centroids are pending. On the application side, the presented tool will be used for a quantitative correlation between individual ply thickness and initiation of transverse cracks as well as spatial variability in interleaf thickness and delamination. 
\vspace{1cm}

\section*{Author contributions}
J.N. developed, implemented and evaluated the approach. J.P.A. provided and described the evaluation micrographs. J.P.A. (\cref{sec: fvf}), C.G. (\cref{sec: interleaf}), and J.B. (\cref{sec: ply thickness}) provided and described the use cases. All authors have read and agreed to the published version of the manuscript. T.d.W. and C.B. supervised the project.
\label{sec: author contributions}

\section*{Algorithms and python packages}
\label{sec packages}
For calculating the bipartite matchings, we are using the \textit{linear\_sum\_assignment()}-function from \textsc{scipy} \cite{Virtanen.2020}.
For the shortest path calculation, we use \textsc{scipy}'s implementation \cite{Virtanen.2020} of the Dijkstra algorithm \cite{Dijkstra.1959}. All EDT are calculated with \textsc{opencv} \cite{Bradski.2000}. Otsu's method is applied using \textsc{scikit-image}'s implementation \cite{vanderWalt.2014}. The $\alpha$-shapes required for the ply boundaries are calculated with the \textsc{alphashape} python package \cite{KenBellock.2021}.

\section*{Code availability statement}
Publication of the code is planned. Further information will be available here in future versions.
\label{sec: code availability}

\bibliographystyle{amsalpha}
\bibliography{ply}

\end{document}